\documentclass{article}

\usepackage{microtype}
\usepackage{graphicx}
\usepackage{subfigure}
\graphicspath{{./images/}} 
\usepackage{booktabs} 

\usepackage{hyperref}

\usepackage[accepted]{icml2025}

\usepackage{amsmath}
\usepackage{amssymb}
\usepackage{mathtools}
\usepackage{amsthm}
\usepackage{bm}
\usepackage{multirow}
\usepackage{textgreek} 

\usepackage[capitalize,noabbrev]{cleveref}

\theoremstyle{plain}
\newtheorem{theorem}{Theorem}[section]

\theoremstyle{definition}

\newtheorem{assumption}[theorem]{Assumption}
\theoremstyle{remark}

\usepackage[textsize=tiny]{todonotes}

\icmltitlerunning{Submission and Formatting Instructions for ICML 2025}

\begin{document}

\twocolumn[
\icmltitle{Information Entropy Invariance: \\
Enhancing Length Extrapolation in Attention Mechanisms}


\icmlsetsymbol{equal}{*}

\begin{icmlauthorlist}
\icmlauthor{Kewei Li}{equal,sch}
\icmlauthor{Yanwen Kong}{equal,sch}
\icmlauthor{Yiping Xu}{sch}
\icmlauthor{Jianlin Su}{yyy}
\icmlauthor{Lan Huang}{sch}
\icmlauthor{Ruochi Zhang}{sch}
\icmlauthor{Fengfeng Zhou}{sch}

\end{icmlauthorlist}

\icmlaffiliation{sch}{College of Computer Science and Technology, and Key Laboratory of Symbolic Computation and Knowledge Engineering of Ministry of Education, Jilin University, Changchun, Jilin, China, 130012}
\icmlaffiliation{yyy}{Moonshot AI Ltd., Beijing, China, 100086}

\icmlcorrespondingauthor{Jianlin Su}{bojone@spaces.ac.cn}
\icmlcorrespondingauthor{Fengfeng Zhou}{FengfengZhou@gmail.com or ffzhou@jlu.edu.cn}

\icmlkeywords{Information Entropy Invariance; Length Extrapolation; Scaled Temperatures; Cosine Attention; Attention Score Dilution.}

\vskip 0.3in
]



\printAffiliationsAndNotice{\icmlEqualContribution} 

\begin{abstract}
Since the emergence of research on improving the length extrapolation capabilities of large language models in 2021, some studies have made modifications to the scaling factor in the scaled dot-product attention mechanism as part of their proposed methods without rigorous theoretical justifications. To fill this gap, we propose two new scaled temperatures based on information entropy invariance to enhance length extrapolation. First, a training-free method InfoScale is designed for dot-product attention, and preserves focus on original tokens during length extrapolation by ensuring consistent entropy. Second, we theoretically analyze the impact of scaling (CosScale) on cosine attention. Experimental data demonstrates that combining InfoScale and CosScale achieves state-of-the-art performance on the GAU-α model with a context window extended to 64 times the training length, and outperforms seven existing methods. Our analysis reveals that significantly increasing CosScale approximates the Windowed Attention, and highlights the significance of attention score dilution as a key challenge in long-range context handling. The code and data are available at \url{https://github.com/HT-NEKO/Information-Entropy-Invariance}.
\end{abstract}

\section{Introduction}
\label{Introduction}

Transformer-based Large Language Models (LLMs) \cite{vaswani2017attention} have achieved remarkable success in various Natural Language Processing (NLP) tasks, particularly those requiring long-range context understanding. However, the limited-length context window imposed by the training data remains a significant bottleneck. This limitation arises partly from the dilution of attention scores by newly added tokens, which disrupts the original distribution and hinders effective long-range modeling \cite{kexuefm-8823}.

While some researchers have explored scaled temperatures to address this issue, a systematic theoretical understanding is lacking. For example, \cite{chiang2022overcoming} proposed a log-length scaler to mitigate the premature plateauing of attention weights observed when training on shorter sequences. Similarly, YaRN \cite{peng2023yarn}  combined pre-softmax scaling with the "NTK-by-part" method, and demonstrated the universal impact of temperature scaling on perplexity across different data samples and token positions within an extended context window. 

This paper introduces a novel approach to improve length extrapolation in LLMs from an information entropy invariance perspective. We begin by investigating scaled dot-product attention and propose a training-free scaling method \textbf{InfoScale} derived from the theoretical foundation established in \cite{kexuefm-8823}. InfoScale aims to maintain focus on original tokens during length extrapolation by preserving information entropy.

To further enhance performance, we extend our analysis to cosine attention. By integrating InfoScale with cosine attention, we achieve significant improvements over existing length extrapolation techniques, including RoPE-based methods (PI \cite{chen2023extending}, YaRN \cite{peng2023yarn}, CLEX \cite{chen2023clex}, ReRoPE \cite{rerope2023}), bias methods (ALiBi \cite{press2021train}), window-based methods (StreamingLLM \cite{xiao2023efficient}, LM-Infinite \cite{han2024lm}, and the Windowed Attention), and skip-wise training methods (PoSE \cite{zhu2023pose}).

Furthermore, we theoretically analyze the behavior of scaled cosine attention (\textbf{CosScale}) and validate these theoretical findings with empirical evidence. We demonstrate that as CosScale increases, the attention mechanism approximates the Windowed Attention. This observation suggests that attention score dilution poses a more significant challenge than insufficient training for extending the context window.

\section{Background and Related Work}
\label{Background and Related Work}
\subsection{Scaled Dot-Product Attention}
Our work builds upon the foundation of scaled dot-product attention introduced in \cite{vaswani2017attention}. Following the same notations, we consider a hidden layer with a set of neurons denoted as $d_{model}$. Given an input sequence of vectors $\boldsymbol{x}_{1},\ldots,~\boldsymbol{x}_{n} \in \boldsymbol{R}^{d_{model}}$,  the attention layer first transforms these vectors into query ($\boldsymbol{Q}$), key ($\boldsymbol{K}$), and value ($\boldsymbol{V}$) vectors:
\begin{equation}
\begin{split}
\bm{Q} &= \bm{[} {\boldsymbol{q}}_{1}, {\boldsymbol{q}}_{2}, {\boldsymbol{q}}_{i}, \ldots, {\boldsymbol{q}}_{n} \bm{]}, ~ {\boldsymbol{q}}_{i} = f_{q} \left( {\boldsymbol{x}}_{i}, i \right) \in \bm{R}^{d_{q}} \\
\bm{K} &= \bm{[} {\boldsymbol{k}}_{1}, {\boldsymbol{k}}_{2}, {\boldsymbol{k}}_{i}, \ldots, {\boldsymbol{k}}_{n} \bm{]}, ~ {\boldsymbol{k}}_{i} = f_{k} \left( {\boldsymbol{x}}_{i}, i \right) \in \bm{R}^{d_{k}} \\
\bm{V} &= \bm{[} {\boldsymbol{v}}_{1}, {\boldsymbol{v}}_{2}, {\boldsymbol{v}}_{i}, \ldots, {\boldsymbol{v}}_{n} \bm{]}, ~ {\boldsymbol{v}}_{i} = f_{v} \left( {\boldsymbol{x}}_{i}, i \right) \in \mathbf{R}^{d_{v}}
\end{split}
\end{equation}
where $d_{model} = H \times d_{q} = H \times d_{k} = H \times d_{v}$, and $H$ is the number of attention heads.

Next, the attention weights are calculated as:
\begin{equation}
\label{eq2}
Attention\left( {\boldsymbol{Q},\boldsymbol{K},\boldsymbol{V}} \right)_{i} = softmax\left( \frac{\boldsymbol{q}_{i}\boldsymbol{K}^{T}}{\sqrt{d_{k}}} \right)\boldsymbol{V}
\end{equation}
where for every vector in $\boldsymbol{K}$, $\boldsymbol{q}_{i}$ and $\boldsymbol{k}_{j}$ are considered as column vectors so that $\boldsymbol{q}_{i}\boldsymbol{k}_{j}^{T}$ is simply the Euclidean inner product. Since softmax will transform the summarization of $\boldsymbol{q}_{i}\boldsymbol{K}^{T}$ to 1, we rewrite the function as
\begin{equation}
\label{eq3}
\begin{split}
{Attention\left( {\boldsymbol{Q},\boldsymbol{K},\boldsymbol{V}} \right)}_{i} = {\sum\limits_{j = 1}^{n}{\beta_{ij}\boldsymbol{v}_{j}}}
\end{split}
\end{equation}
where ${\sum_{j = 1}^{n}\beta_{ij}}=1$, $\beta_{ij}\in\lbrack 0,1\rbrack$ and $\beta_{ij}=\frac{e^{\lambda\boldsymbol{q}_{i}\boldsymbol{k}_{j}^{T}}}{\sum\limits_{j = 1}^{n}e^{\lambda\boldsymbol{q}_{i}\boldsymbol{k}_{j}^{T}}}$.

If $\lambda \hspace{-1mm}=\hspace{-1mm} \frac{1}{\sqrt{d_{k}}}$, the scale temperature used in the vanilla Transformer \cite{vaswani2017attention} plays a crucial role in preventing the dot products from growing too large, which can lead to instability during training.
\subsection{Information Entropy}
Information entropy is a fundamental concept in information theory introduced by Shannon \cite{shannon1948mathematical}. It quantifies the uncertainty or unpredictability inherent in a random variable or an information source. Mathematically, the entropy $~H(\boldsymbol X)$ of a discrete random variable $\boldsymbol X$ is defined as:
\begin{equation}
H\left( \boldsymbol{X} \right) = - {\sum_{i = 1}^{n}{p\left( x_{i} \right){\log{p\left( x_{i} \right)}}}}
\end{equation}
where $p\left( x_{i} \right)$ represents the probability of the $i^{th}$ possible outcome of $\boldsymbol{X}$, and the summation is taken over all possible outcomes. The entropy serves as a measure of the average uncertainty associated with the random variable, and a higher value indicates greater uncertainty. 

Since the softmax operation ensures that ${\sum_{j = 1}^{n}\beta_{ij}} = 1$, $\beta_{ij} \in \lbrack 0,1\rbrack$ in Eq. \ref{eq3}, the outputs of the softmax function adhere to the properties of a probability distribution. We can interpret the attention scores as probability values within the formula for information entropy. This connection enables a theoretical link between attention mechanisms and information entropy, and offers a novel perspective for further analysis \cite{kexuefm-8823}. Then we have the information entropy of the $i^{th}$ token:
\begin{equation}
\label{eq5}
\begin{split}
H_{i} &= - {\sum\limits_{j = 1}^{n}{\beta_{ij}{\log\left( \beta_{ij} \right)}}} \\ &= - \frac{\sum\limits_{j = 1}^{n}{\lambda\boldsymbol{q}_{i}\boldsymbol{k}_{j}^{T}e^{\lambda\boldsymbol{q}_{i}\boldsymbol{k}_{j}^{T}}}}{\sum\limits_{j = 1}^{n}e^{\lambda\boldsymbol{q}_{i}\boldsymbol{k}_{j}^{T}}} + {\mathit{\log}\left( {\sum\limits_{j = 1}^{n}e^{\lambda\boldsymbol{q}_{i}\boldsymbol{k}_{j}^{T}}} \right)}
\end{split}
\end{equation}
where $\beta_{ij}$ denotes the attention score between the $i^{th}$ query token and the $j^{th}$ key token, $\boldsymbol{q}_{i}$ and $\boldsymbol{k}_{j}$  represent the query and key vectors for the $i^{th}$ query token and the $j^{th}$ key token, respectively, and $\lambda$ is a scaling factor that controls the sharpness of the attention distribution. The first term captures the contribution of the individual attention scores, and we rewrite the equation by replacing  $\beta_{ij}$ with $\frac{e^{\lambda \boldsymbol{q}_{i}\boldsymbol{k}_{j}^{T}}}{\sum\limits_{j = 1}^{n}e^{\lambda \boldsymbol{q}_{i}\boldsymbol{k}_{j}^{T}}}$. The resulting entropy provides a quantitative measure of how spread out the attention is across different tokens in the sequence.
\subsection{Scaled Cosine Attention}
Scaled cosine attention is introduced to cooperate with information entropy. Eq. \ref{eq2} can also be rewritten as
\begin{equation}
\label{eq6}
\begin{split}
softmax\left( \frac{\boldsymbol{q}_{i}\boldsymbol{k}_{j}^{T}}{\sqrt{d_{k}}} \right) &= softmax\left( \frac{\left| \left| \boldsymbol{q}_{i} \right| \right|\left| \left| \boldsymbol{k}_{j} \right| \right|{\cos\theta_{ij}}}{\sqrt{d_{k}}} \right) \\&= softmax\left( \frac{\boldsymbol{S}_{ij}~{\cos\theta_{ij}}}{\sqrt{d_{k}}} \right) 
\end{split}
\end{equation}
where $\boldsymbol{S}_{ij}$ is the vector norm product of $\boldsymbol{q}_{i}$ and $\boldsymbol{k}_{j}$, and $\cos\theta_{ij}$ is the cosine of the angle between the two vectors, $\boldsymbol{q}_{i}$ and $\boldsymbol{k}_{j}$, also known as the cosine similarity.
\begin{equation}
\label{eq7}
{\cos\theta_{ij}} = \widehat{\boldsymbol{q}_{i}}~\widehat{\boldsymbol{k}_{j}^{T}} = \frac{\boldsymbol{q}_{i}\boldsymbol{k}_{j}^{T}}{\left| \left| \boldsymbol{q}_{i} \right| \right|\left| \left| \boldsymbol{k}_{j} \right| \right|}
\end{equation}
If we define $\boldsymbol{S}_{ij}$ as a constant $\boldsymbol{S}$ for any pair of column vectors during training, this is referred to as scaled cosine attention. The choice of $\boldsymbol{S}$ is delicate. Making it too small can lead to vanishing gradients, while setting it too large may cause overly sharp attention scores. The latter, however, can be advantageous for tasks involving length extrapolation.
\subsection{Related Work}
\label{3 Related Work}
\textbf{Vanilla Transformer:} The original Transformer \cite{vaswani2017attention} employed a scaling factor of  $\dfrac{1}{\sqrt{d_{k}}}$ in the scaled dot-product attention to mitigate small gradients during backpropagation through the softmax function. However, this design was not specifically intended for tasks requiring length extrapolation.

\textbf{Log-length Scaling:} The method proposed by \cite{chiang2022overcoming} introduces a log-length scaled temperature to address the issue of attention weights plateauing prematurely and attaining lower values when training on shorter sequences. However, this scaling technique tends to fail for very long sequences, and thus, we do not consider it for comparison in our work.

\textbf{Softmax Plus:} \cite{gau-alpha}  extended the log-length scaling idea with an improved temperature scaling method Softmax Plus in the pre-training of a Chinese GAU-\textalpha~  model. This method attempts to provide a theoretical explanation for the effectiveness of log-length scaling, grounded in the concept of information entropy invariance. However, it has not been systematically analyzed in depth.

\textbf{Pre-softmax Scaling:} The pre-softmax scaling approach proposed by \cite{peng2023yarn} for the YaRN model experimentally showed that, for an appropriate scaling factor, the model could achieve better perplexity scores across extended context windows. However, this method is tightly integrated with YaRN, which dynamically modifies the frequency of each dimension of RoPE during inference. As our method is independent of RoPE, we exclude it from the comparison. 

\textbf{Cosine Similarity:} In the field of face recognition, several studies have explored the use of cosine similarity loss derived from softmax \cite{wang2018additive, deng2019arcface, wang2018cosface} The scale of the cosine similarity was typically set to fixed values such as 30 \cite{wang2018additive}, 64 \cite{deng2019arcface}, or adaptively adjusted based on a lower bound in multiclass classification tasks \cite{wang2018cosface}. It is generally understood that the scale should not be too small or too large, with an optimal range around 50. For length extrapolation tasks, however, using a larger scaling factor in the scaled cosine attention has been shown to yield better performance on longer context windows, as it results in a more concentrated distribution of attention scores. We present two theorems related to CosScale in \cref{Theorems of the Behaviors of CosScale}.

Please see \cref{table1} in \cref{Additional Tables and Charts} for details of each method.

\section{Methodology}
\label{Methodology}
We firstly analyze the information entropy invariance of a novel scaled temperature InfoScale for dot-product attention in \cref{Theoretical Foundation of InfoScale}. Motivated by the assumptions underlying InfoScale, we turn our attention to cosine attention in \cref{Theorems of the Behaviors of CosScale}. Here, we establish two theorems that characterize the behavior of its scaled temperature, which we refer to as CosScale. These theorems provide a theoretical framework for understanding the impact of CosScale on length extrapolation.
\subsection{Theoretical Foundation of InfoScale}
\label{Theoretical Foundation of InfoScale}
Building on the foundation theory by \cite{shannon1948mathematical}, we continue the expansion along Eq. \ref{eq5} and introduce three assumptions for further deduction (See \cref{The details of the deduction of InfoScale} for the details).
\begin{assumption}
\label{ass1}
The attention scores follow the Law of Large Numbers \cite{kolmogorov2018foundations}, which implies ${\sum_{i = 1}^{n}y_{i}} = n \times \frac{1}{n}{\sum_{i = 1}^{n}y_{i}} = n\mathbb{E}\left\lbrack y_{i} \right\rbrack$.
\end{assumption}
Under \cref{ass1}, we can approximate the function in Eq. \ref{eq5} as follows:
\begin{equation}
\label{eq8}
H_{i} \cong \widehat{H_{i}} = {\log(n)} + {\log\left( {\mathbb{E}\left\lbrack e^{\lambda \boldsymbol{q}_{i}\boldsymbol{k}_{j}^{T}} \right\rbrack} \right)} - \frac{\mathbb{E}\left\lbrack {\lambda \boldsymbol{q}_{i}\boldsymbol{k}_{j}^{T}e^{\lambda \boldsymbol{q}_{i}\boldsymbol{k}_{j}^{T}}} \right\rbrack}{\mathbb{E}\left\lbrack e^{\lambda \boldsymbol{q}_{i}\boldsymbol{k}_{j}} \right\rbrack}
\end{equation}
Here,  $\widehat{H_{i}}$ represents the approximation of $H_{i}$ using the Law of Large Numbers \cite{kolmogorov2018foundations}. We now need to calculate two expectations $\mathbb{E}\left\lbrack e^{\lambda \boldsymbol{q}_{i}\boldsymbol{k}_{j}^{T}} \right\rbrack$ and $\mathbb{E}\left\lbrack {\lambda \boldsymbol{q}_{i}\boldsymbol{k}_{j}^{T}e^{\lambda \boldsymbol{q}_{i}\boldsymbol{k}_{j}^{T}}} \right\rbrack$, which are not directly computable. Fortunately, by applying \cref{ass2}, which is informed by the weight initialization strategy \cite{glorot2010understanding}, we can proceed with further derivations.

\begin{assumption}
\label{ass2}
The word embeddings lie within a $d_{k}$-dimensional hypersphere with radius $r = \sqrt{vd_{k}}$, where $v$ represents the variance of the word embeddings during initialization. In this case, the expectation of $\boldsymbol{q}_{i}\boldsymbol{k}_{j}^{T}$ corresponds to $r^2\cos\theta$, where $\theta$ is the angle between $\boldsymbol{q}_{i}$ and $\boldsymbol{k}_{j}$. With this assumption, we can express $\lambda \boldsymbol{q}_{i}\boldsymbol{k}_{j}^{T} = \lambda r^{2}\cos\theta = \lambda vd_{k}\cos\theta = \alpha~\cos\theta$.\end{assumption} Substituting this into the expression of $\widehat{H_{i}}$, we obtain:
\begin{equation}
\label{eq9}
\begin{split}
\widehat{H_{i}} = ~&{\mathit{\log}(n)} + {\log\left( \frac{\int_{0}^{\pi}{e^{\alpha \cos\theta}{\mathit{\sin}^{d_{k} - 2}\theta}~\mathrm{d}\theta}}{\int_{0}^{\pi}{{\mathit{\sin}^{d_{k} - 2}\theta}~\mathrm{d}\theta}} \right)} \\
&- \frac{\int_{0}^{\pi}{\alpha \cos\theta e^{\alpha \cos\theta}{\mathit{\sin}^{d_{k} - 2}\theta}~\mathrm{d}\theta}}{\int_{0}^{\pi}{e^{\alpha \cos\theta}{\mathit{\sin}^{d_{k} - 2}\theta}~\mathrm{d}\theta}}
\end{split}
\end{equation}
To solve this equation with three integrals, we introduce \cref{ass3}. 
\begin{assumption}
\label{ass3}
 The integrals can be approximated using Laplace’s approximation \cite{kruschke2010bayesian}.
\end{assumption}
Applying this approximation, we arrive at the following expression:
\begin{equation}
\label{eq10}
\widehat{H_{i}} \cong {\mathit{\log}(n)} + \frac{d}{2}\left\lbrack \left. {\mathit{\log}\left( \frac{\left( {\sqrt{4t + 1} - 1} \right)}{2t} \right.} \right) \right\rbrack,~t = \lambda^{2}v^{2}
\end{equation}
As $\left. \lambda\rightarrow 0 \right.$ and $
\left. v\rightarrow 0 \right.$ (See \cref{The details of the deduction of InfoScale} for the details), we perform a Taylor expansion \cite{stewart2012calculus} of $\left. \sqrt{4t + 1} - 1 \right)$ at $t = 0$:
\begin{equation}
\label{eq11}
\sqrt{4t + 1} - 1 = 2t - 2t^{2} + o\left( t^{3} \right)
\end{equation}
Substituting Eq. \ref{eq11} into Eq. \ref{eq10}, we get:
\begin{equation}
\label{eq12}
\begin{split}
\widehat{H_{i}} &\cong {\mathit{\log}(n)} + \frac{d_{k}}{2}\left\lbrack \left. {\mathit{\log}\left( \frac{\left( {2t - 2t^{2} + o\left( t^{3} \right)} \right)}{2t} \right.} \right) \right\rbrack \\
&\cong {\mathit{\log}(n)} + \frac{d_{k}}{2}\left\lbrack \left. {\mathit{\log}\left( 1 - t \right.} \right) \right\rbrack = \epsilon
\end{split}
\end{equation}
where $\epsilon$ is a hyperparameter that ensures the information entropy of the attention scores remains constant regardless the input length $n$.
\begin{equation}
\label{eq13}
\lambda = \frac{\sqrt{1 - e^{\frac{2\epsilon}{d_{k}}}n^{- \frac{2}{d_{k}}}}}{v} = \mathcal{K}\sqrt{1 - e^{\frac{2\epsilon}{d_{k}}}n^{- \frac{2}{d_{k}}}}
\end{equation}
When the model is pre-trained by a maximum length $n_{tr}$ with the scaling factor $\frac{1}{\sqrt{d_{k}}}$ and $n = n_{tr}$, we assume the model has been perfectly trained. Thus, we have:
\begin{equation}
\label{eq14}
\begin{split}
\frac{1}{\sqrt{d_{k}}} &= \lambda = \mathcal{K}\sqrt{1 - e^{\frac{2\epsilon}{d_{k}}}{n_{tr}}^{- \frac{2}{d_{k}}}},\\\mathcal{K} &= \frac{1}{\sqrt{\left( {1 - e^{\frac{2\epsilon}{d_{k}}}{n_{tr}}^{- \frac{2}{d_{k}}}} \right)d_{k}}}
\end{split}
\end{equation}
Substituting the expression for $\mathcal{K}$ into Eq. \ref{eq13}, we obtain:
\begin{equation}
\begin{split}
\lambda &= \frac{\sqrt{1 - e^{\frac{2\epsilon}{d_{k}}}n^{- \frac{2}{d_{k}}}}}{\sqrt{\left( {1 - e^{\frac{2\epsilon}{d_{k}}}{n_{tr}}^{- \frac{2}{d_{k}}}} \right)d_{k}}} \\
&= \frac{\sqrt{1 - e^{\frac{2\epsilon}{d_{k}}}n^{- \frac{2}{d_{k}}}}}{\sqrt{\left( {1 - e^{\frac{2\epsilon}{d_{k}}}{n_{tr}}^{- \frac{2}{d_{k}}}} \right)}}*\frac{1}{\sqrt{d_{k}}} = \frac{InfoScale}{\sqrt{d_{k}}}
\end{split}
\end{equation}
Thus, we define the scaled temperature $\displaystyle\sqrt{\frac{1 - e^{\frac{2\epsilon}{d_{k}}}n^{- \frac{2}{d_{k}}}}{{1 - e^{\frac{2\epsilon}{d_{k}}}{n_{tr}}^{- \frac{2}{d_{k}}}}}}$ as InfoScale.
\subsection{Theorems of the Behaviors of CosScale}
\label{Theorems of the Behaviors of CosScale}
In \cref{Theoretical Foundation of InfoScale}, we discussed \cref{ass2} for the derivations of InfoScale. However, in practical situations, the conditions may not fully align with this assumption after training. Specifically, the word embeddings of $q_{i}$ and $k_{j}$ may not lie on a $d_{k}$-dimensional rational hypersphere. To address this issue, we introduce CosScale in conjunction with cosine attention to enforce the model’s alignment with \cref{ass2}. 
 In this section, we present two theorems regarding the behavior of CosScale, the scaling factor in scaled cosine attention, which will be referred to as $\alpha$ for the purpose of deduction.
(See \cref{The details of the deduction of CosScale} for detailed derivations and \cref{Testify the validation of the CosScale's Theorems} for experimental validation results.)
\begin{theorem}
\label{theo1}
 The peak value $\eta_{1}^{*}$ of the $QK$ distribution before RoPE shifts toward 1 as CosScale $\alpha$ increases.
\end{theorem}
Given that the GAU-α model applies a RoPE operation to $QK$ before the softmax normalization, we have $a_{ij} = e^{\alpha{\cos\theta}_{ij}}$. Then we got its expectation:
\begin{equation}
\label{eq16}
\mathbb{E}\left\lbrack a_{ij} \right\rbrack = \frac{1}{n^{2}}{\sum_{ij}^{n^{2}}e^{\raisebox{0.6ex}{\scalebox{0.8}{$\frac{\alpha \boldsymbol{q}_{i}\boldsymbol{k}_{j}^{T}}{{|{|\boldsymbol{q}_{i}|}|}{|{|\boldsymbol{k}_{j}|}|}}$}}}} = \frac{1}{n^{2}}{\sum_{ij}^{n^{2}}e^{\alpha{\cos\theta}_{ij}}} \cong \mathbb{E}\left\lbrack e^{\alpha \cos\theta} \right\rbrack
\end{equation}
Under the assumption from \cref{ass2}, where $\eta = \cos\theta$, we can express this expectation as:
\begin{equation}
\label{eq17}
\mathbb{E}\left\lbrack a_{ij} \right\rbrack \cong \mathbb{E}\left\lbrack e^{\alpha \cos\theta} \right\rbrack = \frac{{\int_{- 1}^{1}{e^{\alpha\eta}\left( {1 - \eta^{2}} \right)^{\frac{d_{k} - 3}{2}}}}\mathrm{d}\eta}{{\int_{- 1}^{1}\left( {1 - \eta^{2}} \right)^{\frac{d_{k} - 3}{2}}}\mathrm{d}\eta} = \frac{\boldsymbol{{I}_{1}}}{\boldsymbol{{I}_{0}}}
\end{equation}
Here, the maximum value of $\boldsymbol{{I}_{1}}$ occurs at
\begin{equation}
\label{eq18}
\eta_{1}^{*} = \frac{- \left( {d_{k} - 3} \right) + \sqrt{\left( {d_{k} - 3} \right)^{2} + 4\alpha^{2}}}{2\alpha}
\end{equation}
It follows that the exponential functions $e^{\alpha\eta}$ causes the integrals $\boldsymbol{{I}_{1}}$ to reach their maximum value as $\left. \eta\rightarrow 1 \right.$ when $\alpha$ becomes larger. \cref{f7} illustrates the relationship between $\alpha$ and $\eta_{1}^{*}$, \textit{i.e.}, as $\alpha$ increases, $\eta_{1}^{*}$ approaches 1.
\begin{theorem}
RoPE becomes more dominant in attention score as $\alpha$ increases.
\end{theorem}
Based on \cref{theo1}, as $\alpha$ increases, the values of $\boldsymbol{q}_{i}\boldsymbol{k}_{j}$ for all $j \in \lbrack 1,n\rbrack$ and any token $x_{i}$ converge. It leads to the dominance of RoPE in distinguishing attention scores between different tokens.
Specifically, for any token $x_{i}$, the attention scores $\boldsymbol{q}_{i}\boldsymbol{k}_{j}$ become increasingly similar for all $j$, making RoPE the primary factor in differentiating the attention across tokens. This phenomenon emphasizes the growing importance of RoPE as $\alpha$ increases.

Note that CosScale is a scaling factor only used in scaled cosine attention. In the following sections, we will refer to CosScale without repeatedly emphasizing that it operates within the context of scaled cosine attention.
\section{Experiments}
\label{Experiments}
This section evaluates the effectiveness of InfoScale and CosScale in handling long contexts and improving the length extrapolation capabilities of LLMs. We benchmark our proposed methods against a diverse set of baselines, including: \textbf{RoPE-based methods} (PI \cite{chen2023extending}, YaRN \cite{peng2023yarn},  ReRoPE \cite{rerope2023}), \textbf{Bias methods} (ALiBi \cite{press2021train}), \textbf{Window-based methods} (StreamingLLM \cite{xiao2023efficient}, LM-Infinite \cite{han2024lm}, the Windowed Attention), and \textbf{Skip-wise training method} (PoSE \cite{zhu2023pose}). Implementation details for each method are provided in \cref{Model Training and Fine-tuning Process}. We assess the performance of InfoScale and CosScale using two metrics: perplexity (PPL) and masked-token prediction accuracy (ACC), whose detailed definitions are in \cref{Evaluation Method}. Evaluation is conducted on sequences up to 64 times longer than the maximum sequence length encountered during training, allowing us to rigorously assess length extrapolation capabilities.
\subsection{Dataset and Implementation Details}
\subsubsection{Dataset}
We utilize the Wanjuan Patent Dataset (WJ), a subset of the LongData-Corpus dataset \cite{yu2023paraphrasing} specifically chosen for its suitability in evaluating long-range context handling. The LongData-Corpus dataset comprises samples with an average length exceeding 16k tokens. To ensure our experiments focus on long sequences, the WJ dataset in this study includes only samples with a text length greater than 4k tokens. Detailed information about the WJ dataset and its partitioning is provided in \cref{Dataset}.
\subsubsection{Experiment Settings}
\textbf{Training Procedure:} For each experiment, we randomly mask 15\% of the tokens in each training sample. We employ the GAU-α architecture with the top 6 layers \cite{gau-alpha}. Due to resource constraints, we limit our evaluation to GAU-α and do not include experiments with larger models like LLaMA \cite{touvron2023llama}, as we are unable to perform inference with sequences longer than 1.5 times the maximum training length on a 7B parameter model (the smallest LLaMA variant).

The training process consists of 400 epochs with a warm-up period comprising 10\% of the total training steps. We use a linear learning rate scheduler with an initial learning rate of 2e-5. The maximum sequence length during training is 64, and a global batch size of 128 is distributed across 4 Tesla V100-SXM2-32GB GPUs. Optimization is performed using the AdamW optimizer with default hyperparameter values. Details on the model training loss can be found in \cref{Model Training and Fine-tuning Process}.

\textbf{Fine-tuning Procedure:} For baseline methods that require fine-tuning, we perform an additional fine-tuning step for 100 epochs and keep all other configurations consistent with the corresponding training procedure of the method. See \cref{Fine-tuning Process} in detail.

\textbf{Evaluation Procedure:} We evaluate models using a token-level masking approach, where the model predicts the masked token for validation. A global batch size of 32 is used. The hyperparameter $\epsilon$ for InfoScale is set to 0.

\subsection{Experimental Results}
It is important to note that for sequence lengths shorter than 512, some baseline algorithms may perform slightly better than our methods (InfoScale, CosScale, or their combination). This is likely because the original GAU-α model was trained with a maximum sequence length of 512, enabling it to capture positional information more effectively within this range. However, as we extrapolate to longer sequences (\textit{e.g.}, 4k), the benefits of our proposed methods become evident.
\subsubsection{InfoScale Demonstrates Length Extrapolation Capabilities}
When evaluated on sequences of length 4k, InfoScale significantly improves the performance of several baseline methods. For instance, InfoScale boosts the accuracy (ACC) of ReRoPE \cite{rerope2023} by nearly tenfold while simultaneously halving its perplexity (PPL) (see \cref{table 4.2.3} and \cref{table 4.2.1}). While InfoScale consistently improves ACC for most baselines, LM-Infinite and the Windowed Attention (W.A.) exhibit a slight increase in PPL despite improved ACC. This discrepancy may be attributed to the assumption in InfoScale's derivation that high-dimensional word embeddings are distributed on a fixed-radius hypersphere, an assumption not strictly satisfied by scaled dot-product attention. To address this, we introduce cosine attention, which enforces this constraint on the embedding distribution. (A detailed discussion of the results is provided in \cref{All published methods fail on GAU-α}. )

\subsubsection{CosScale Demonstrates Strong Length Extrapolation Capabilities}
As shown in \cref{table 4.2.3} and \cref{table 4.2.2}, when evaluated on sequences of length 4k, CosScale consistently improves the performance of most baseline models. With the exception of PI and ALiBi whose poor performance is attributed to limitations inherent to the baselines themselves, all other methods exhibit gains when incorporating CosScale. The improvements are particularly pronounced for ReRoPE and PoSE. CosScale enhances the ACC of ReRoPE by nearly tenfold and reduces its PPL by over a hundred times. Similarly, PoSE achieves a nearly sixteen-fold improvement in accuracy and a reduction in perplexity by almost 27 times. These results demonstrate the substantial performance gains achieved with CosScale.

It is worth noting that we set the CosScale for the Windowed Attention to 16 \cite{kexuefm-9812}, rather than 128 as used for other algorithms. This is because the context window for the Windowed Attention is 64, which falls outside the long-range context regime where CosScale excels.

These findings suggest that constraining the distribution of word embeddings to a hypersphere facilitates more effective angular training of the attention matrix. Based on these results, CosScale alone exhibits strong length extrapolation capabilities. Furthermore, since InfoScale assumes that the word embeddings are distributed on a hypersphere, combining CosScale with InfoScale is expected to yield even greater performance improvements.
\subsubsection{Combining InfoScale and CosScale for Enhanced Performance}
As shown in \cref{table 4.2.3}, the combined application of CosScale and InfoScale leads to further performance improvements compared to using either method alone, particularly when extrapolating to a sequence length of 4k. For instance, the ACC improvement for PoSE increases from 15.699x with CosScale alone to 16.957x with both CosScale and InfoScale. Similarly, for ReRoPE, the improvement rises from 10.771x to 11.371x. This synergistic effect suggests that the combination of CosScale and InfoScale results in a performance boost greater than the sum of their individual contributions. These findings align with our hypothesis that constraining the word embeddings to a hypersphere (CosScale) enhances the effectiveness of InfoScale, which is derived under the same assumption.
\begin{table*}[]
\caption{Perplexity (PPL) and accuracy (ACC) for baseline methods without or with both CosScale and InfoScale incorporated, evaluated with sequence lengths ranging from 128 to 4k on the length extrapolation task. Lower PPL and higher ACC indicate better performance. CosScale is set to 128 for all methods except the Windowed Attention, where it is set to 16. W.A. denotes the Windowed Attention, while I.S. and C.S. represent InfoScale and CosScale, respectively.}
\label{table 4.2.3}
\vskip 0.1in
\begin{center}
\begin{tabular}{l@{\hspace{0.08cm}}|c@{\hspace{0.03cm}}c@{\hspace{0.03cm}}l@{\hspace{0.2cm}}c@{\hspace{0.03cm}}c@{\hspace{0.03cm}}l@{\hspace{0.2cm}}c@{\hspace{0.03cm}}c@{\hspace{0.03cm}}l@{\hspace{0.2cm}}c@{\hspace{0.03cm}}c@{\hspace{0.03cm}}l@{\hspace{0.2cm}}c@{\hspace{0.03cm}}c@{\hspace{0.03cm}}l@{\hspace{0.2cm}}c@{\hspace{0.03cm}}c@{\hspace{0.03cm}}}
\toprule
\multirow{3}{*}{Methods}    & \multicolumn{17}{c}{Evaluation Length}                                                                                                                                                                                                                                          \\ \cline{2-18} 
                            & \multicolumn{2}{c}{128}        &           & \multicolumn{2}{c}{256}        &           & \multicolumn{2}{c}{512}            &           & \multicolumn{2}{c}{1024}           &           & \multicolumn{2}{c}{2048}           &           & \multicolumn{2}{c}{4096}           \\ \cline{2-3} \cline{5-6} \cline{8-9} \cline{11-12} \cline{14-15} \cline{17-18} 
                            & PPL           & ACC.           &           & PPL           & ACC.           &           & PPL               & ACC.           &           & PPL               & ACC.           &           & PPL               & ACC.           &           & PPL               & ACC.           \\ \hline
PI                          & 24.51         & 0.46           &           & 10.75         & 0.55           &           & \textgreater{}500 & \textless{}0.1 &           & \textgreater{}500 & \textless{}0.1 &           & \textgreater{}500 & \textless{}0.1 &           & \textgreater{}500 & \textless{}0.1 \\
PI w/ C.S.                  & 10.38         & 0.52           &           & 10.46         & 0.53           &           & \textgreater{}500 & \textless{}0.1 &           & \textgreater{}500 & \textless{}0.1 &           & \textgreater{}500 & \textless{}0.1 &           & \textgreater{}500 & \textless{}0.1 \\
PI w/ I.S.                  & 21.67         & 0.48           &           & 8.14          & 0.61           &           & \textgreater{}500 & \textless{}0.1 &           & \textgreater{}500 & \textless{}0.1 &           & \textgreater{}500 & \textless{}0.1 &           & \textgreater{}500 & \textless{}0.1 \\
PI w/ C.S., I.S.            & 9.59          & 0.54           &           & 10.15         & 0.54           &           & \textgreater{}500 & \textless{}0.1 &           & \textgreater{}500 & \textless{}0.1 &           & \textgreater{}500 & \textless{}0.1 &           & \textgreater{}500 & \textless{}0.1 \\ \hline
ALiBi                       & 151.00        & \textless{}0.1 &           & 271.97        & \textless{}0.1 &           & 346.88            & \textless{}0.1 &           & 401.87            & \textless{}0.1 &           & 488.47            & \textless{}0.1 &           & \textgreater{}500 & \textless{}0.1 \\
ALiBi w/ C.S.               & 151.57        & \textless{}0.1 &           & 263.90        & \textless{}0.1 &           & 335.35            & \textless{}0.1 &           & 392.22            & \textless{}0.1 &           & 468.78            & \textless{}0.1 &           & 486.49            & \textless{}0.1 \\
ALiBi w/ I.S.               & 150.74        & \textless{}0.1 &           & 270.76        & \textless{}0.1 &           & 353.25            & \textless{}0.1 &           & 404.16            & \textless{}0.1 &           & 478.59            & \textless{}0.1 &           & \textgreater{}500 & \textless{}0.1 \\
ALiBi w/ C.S., I.S.         & 147.23        & 0.11           &           & 258.90        & \textless{}0.1 &           & 322.28            & \textless{}0.1 &           & 386.62            & \textless{}0.1 &           & 444.07            & \textless{}0.1 &           & 473.05            & \textless{}0.1 \\ \hline
PoSE                        & 1.97          & 0.85           &           & 2.23          & 0.81           &           & 2.61              & 0.77           &           & 14.74             & 0.46           &           & 240.25            & 0.10           &           & \textgreater{}500 & \textless{}0.1 \\
PoSE w/ C.S.                & 2.07          & 0.84           &           & 2.29          & 0.81           &           & 2.34              & 0.81           &           & 3.35              & 0.73           &           & 12.37             & 0.51           &           & 22.03             & 0.41           \\
PoSE w/ I.S.                & \textbf{1.95} & \textbf{0.85}  & \textbf{} & \textbf{2.10} & \textbf{0.83}  & \textbf{} & \textbf{2.15}     & \textbf{0.82}  & \textbf{} & 6.32              & 0.60           &           & 73.80             & 0.23           &           & 354.50            & \textless{}0.1 \\
PoSE w/ C.S., I.S.          & 2.06          & 0.84           &           & 2.24          & 0.82           &           & 2.28              & 0.81           &           & \textbf{3.30}     & \textbf{0.74}  & \textbf{} & \textbf{11.20}    & \textbf{0.53}  & \textbf{} & \textbf{18.41}    & \textbf{0.44}  \\ \hline
StreamingLLM                & 2.18          & 0.84           &           & 2.55          & 0.80           &           & 2.85              & 0.78           &           & 4.27              & 0.71           &           & 4.88              & 0.69           &           & 5.42              & 0.67           \\
StreamingLLM w/ C.S.        & 2.31          & 0.82           &           & 2.72          & 0.78           &           & 2.97              & 0.77           &           & 4.13              & 0.72           &           & 4.59              & 0.70           &           & 4.88              & 0.69           \\
StreamingLLM w/ I.S.        & \textbf{2.18} & \textbf{0.83}  & \textbf{} & \textbf{2.53} & \textbf{0.80}  & \textbf{} & \textbf{2.85}     & \textbf{0.79}  & \textbf{} & 4.76              & 0.70           &           & 4.84              & 0.70           &           & 5.20              & 0.68           \\
StreamingLLM w/ C.S., I.S.  & 2.29          & 0.82           &           & 2.68          & 0.79           &           & 2.94              & 0.78           &           & \textbf{4.13}     & \textbf{0.72}  & \textbf{} & \textbf{4.54}     & \textbf{0.71}  & \textbf{} & \textbf{4.83}     & \textbf{0.70}  \\ \hline
YaRN(s=16)                  & 2.41          & 0.81           &           & 2.99          & 0.76           &           & 3.37              & 0.73           &           & 7.07              & 0.60           &           & 15.34             & 0.48           &           & 26.59             & 0.41           \\
YaRN(s=16) w/ C.S.          & 2.57          & 0.78           &           & 3.09          & 0.75           &           & 3.09              & 0.75           &           & 4.79              & 0.67           &           & 6.28              & 0.63           &           & 7.73              & 0.59           \\
YaRN(s=16) w/ I.S.          & \textbf{2.26} & \textbf{0.82}  & \textbf{} & \textbf{2.58} & \textbf{0.80}  & \textbf{} & \textbf{2.61}     & \textbf{0.78}  & \textbf{} & 4.37              & \textbf{0.70}  & \textbf{} & 7.49              & 0.61           &           & 12.38             & 0.53           \\
YaRN(s=16) w/ C.S., I.S.    & 2.47          & 0.79           &           & 2.83          & 0.77           &           & 2.75              & 0.77           &           & \textbf{4.30}     & 0.70           &           & \textbf{5.84}     & \textbf{0.65}  & \textbf{} & \textbf{7.52}     & \textbf{0.60}  \\ \hline
YaRN(s=32)                  & 2.65          & 0.79           &           & 3.35          & 0.74           &           & 3.88              & 0.71           &           & 7.94              & 0.58           &           & 16.68             & 0.47           &           & 27.59             & 0.40           \\
YaRN(s=32) w/ C.S.          & 2.70          & 0.78           &           & 3.29          & 0.73           &           & 3.40              & 0.73           &           & 5.29              & 0.65           &           & 6.87              & 0.61           &           & 8.50              & 0.58           \\
YaRN(s=32) w/ I.S.          & \textbf{2.42} & \textbf{0.81}  & \textbf{} & \textbf{2.74} & \textbf{0.78}  & \textbf{} & \textbf{2.75}     & \textbf{0.77}  & \textbf{} & \textbf{4.40}     & \textbf{0.69}  & \textbf{} & 7.04              & 0.62           &           & 10.56             & 0.55           \\
YaRN(s=32) w/ C.S., I.S.    & 2.57          & 0.79           &           & 2.98          & 0.76           &           & 3.00              & 0.75           &           & 4.68              & 0.68           &           & \textbf{6.26}     & \textbf{0.63}  & \textbf{} & \textbf{8.28}     & \textbf{0.59}  \\ \hline
W.A.                        & 2.19          & 0.84           &           & 2.55          & 0.80           &           & 2.82              & 0.78           &           & 3.69              & 0.74           &           & 3.87              & 0.73           &           & 4.24              & 0.71           \\
W.A.(C.S.=16) w/ C.S.       & 2.21          & 0.84           &           & 2.67          & 0.79           &           & 2.93              & 0.78           &           & 3.81              & 0.73           &           & 4.02              & 0.73           &           & 4.44              & 0.71           \\
W.A. w/ I.S.                & 2.19          & 0.83           &           & 2.53          & 0.80           &           & 2.83              & 0.79           &           & 3.73              & 0.74           &           & 3.92              & 0.74           &           & 4.32              & 0.72           \\
W.A.(C.S.=16) w/ C.S., I.S. & \textbf{2.17} & \textbf{0.84}  & \textbf{} & \textbf{2.51} & \textbf{0.81}  & \textbf{} & \textbf{2.72}     & \textbf{0.80}  & \textbf{} & \textbf{3.54}     & \textbf{0.75}  & \textbf{} & \textbf{3.72}     & \textbf{0.75}  & \textbf{} & \textbf{4.06}     & \textbf{0.73}  \\ \hline
LM-Infinite                 & 12.25         & 0.55           &           & 18.65         & 0.46           &           & 21.66             & 0.43           &           & 30.72             & 0.39           &           & 32.04             & 0.38           &           & 32.56             & 0.37           \\
LM-Infinite w/ C.S.         & 4.09          & 0.69           &           & 6.63          & 0.61           &           & 7.99              & 0.58           &           & 10.88             & 0.54           &           & 11.83             & 0.53           &           & 12.91             & 0.50           \\
LM-Infinite w/ I.S.         & 12.66         & 0.54           &           & 26.64         & 0.42           &           & 28.47             & 0.40           &           & 30.90             & 0.38           &           & 29.92             & 0.38           &           & 32.81             & 0.37           \\
LM-Infinite w/ C.S., I.S.   & \textbf{3.97} & \textbf{0.71}  & \textbf{} & \textbf{6.22} & \textbf{0.63}  & \textbf{} & \textbf{7.26}     & \textbf{0.60}  & \textbf{} & \textbf{9.82}     & \textbf{0.56}  & \textbf{} & \textbf{10.42}    & \textbf{0.56}  & \textbf{} & \textbf{11.11}    & \textbf{0.54}  \\ \hline
ReRoPE                      & 15.61         & 0.49           &           & 46.44         & 0.33           &           & 151.22            & 0.18           &           & 295.42            & 0.12           &           & 475.82            & \textless{}0.1 &           & \textgreater{}500 & \textless{}0.1 \\
ReRoPE w/ C.S.              & 3.69          & 0.72           &           & 3.62          & 0.72           &           & 3.88              & 0.71           &           & 5.10              & 0.67           &           & 5.55              & 0.66           &           & 6.36              & 0.63           \\
ReRoPE w/ I.S.              & 14.77         & 0.50           &           & 23.61         & 0.43           &           & 52.86             & 0.32           &           & 115.83            & 0.23           &           & 210.78            & 0.17           &           & 311.24            & 0.12           \\
ReRoPE w/ C.S., I.S.        & \textbf{3.64} & \textbf{0.72}  & \textbf{} & \textbf{3.47} & \textbf{0.73}  & \textbf{} & \textbf{3.63}     & \textbf{0.73}  & \textbf{} & \textbf{4.74}     & \textbf{0.69}  & \textbf{} & \textbf{4.97}     & \textbf{0.69}  & \textbf{} & \textbf{5.41}     & \textbf{0.67} 
 \\
\bottomrule
\end{tabular}
\end{center}
\end{table*}
\begin{table*}
\caption{Comparison of InfoScale and Softmax Plus (abbreviated as softmax+) with CosScale on length extrapolation performance.}
\label{table 4.2.4-1}
\vskip 0.1in
\centering
    \begin{tabular}{l@{\hspace{0.1cm}}|c@{\hspace{0.1cm}}c@{\hspace{0.1cm}}l@{\hspace{0.25cm}}c@{\hspace{0.1cm}}c@{\hspace{0.1cm}}l@{\hspace{0.25cm}}c@{\hspace{0.1cm}}c@{\hspace{0.1cm}}l@{\hspace{0.25cm}}c@{\hspace{0.1cm}}c@{\hspace{0.1cm}}l@{\hspace{0.25cm}}c@{\hspace{0.1cm}}c@{\hspace{0.1cm}}l@{\hspace{0.25cm}}c@{\hspace{0.1cm}}c@{\hspace{0.1cm}}}
\toprule
\multicolumn{1}{l|}{\multirow{3}{*}{Methods}}  & \multicolumn{17}{c}{Evaluation Length}                                                                                                                                                                                                \\ \cline{2-18} 
\multicolumn{1}{l|}{}                          & \multicolumn{2}{c}{128}       &  & \multicolumn{2}{c}{256}       &  & \multicolumn{2}{c}{512}       &  & \multicolumn{2}{c}{1024}      &  & \multicolumn{2}{c}{2048}        &  & \multicolumn{2}{c}{4096}                             \\ \cline{2-3} \cline{5-6} \cline{8-9} \cline{11-12} \cline{14-15} \cline{17-18} 
\multicolumn{1}{l|}{}                          & PPL           & ACC.          &  & PPL           & ACC.          &  & PPL           & ACC.          &  & PPL           & ACC.          &  & PPL             & ACC.          &  & PPL                        & ACC.                    \\ \hline
\multicolumn{1}{l|}{GAU-α   w/ C.S., I.S.}     & \textbf{2.28} & \textbf{0.82} &  & \textbf{2.64} & \textbf{0.79} &  & \textbf{2.86} & \textbf{0.78} &  & 5.03          & \textbf{0.68} &  & 23.24           & \textbf{0.44} &  & \textbf{44.07}             & \textbf{0.34}           \\
\multicolumn{1}{l|}{GAU-α   w/ C.S., softmax+} & 2.56          & 0.78          &  & 2.82          & 0.77          &  & 2.92          & 0.77          &  & \textbf{4.96} & 0.68          &  & \textbf{23.01}  & 0.44          &  & 44.08                      & 0.34                    \\ \bottomrule
\end{tabular}
\end{table*}
\begin{table*}
\caption{Comparison of InfoScale and Softmax Plus (abbreviated as softmax+) on the baseline without CosScale on length extrapolation performance.}
\label{table 4.2.4-2}
\vskip 0.1in
\centering
    \begin{tabular}{l@{\hspace{0.1cm}}|c@{\hspace{0.1cm}}c@{\hspace{0.1cm}}l@{\hspace{0.25cm}}c@{\hspace{0.1cm}}c@{\hspace{0.1cm}}l@{\hspace{0.25cm}}c@{\hspace{0.1cm}}c@{\hspace{0.1cm}}l@{\hspace{0.25cm}}c@{\hspace{0.1cm}}c@{\hspace{0.1cm}}l@{\hspace{0.25cm}}c@{\hspace{0.1cm}}c@{\hspace{0.1cm}}l@{\hspace{0.25cm}}c@{\hspace{0.1cm}}c@{\hspace{0.1cm}}}
\toprule
\multicolumn{1}{l|}{\multirow{3}{*}{Methods}}  & \multicolumn{17}{c}{Evaluation Length}                                                                                                                                                                                                \\ \cline{2-18} 
\multicolumn{1}{l|}{}                          & \multicolumn{2}{c}{128}       &  & \multicolumn{2}{c}{256}       &  & \multicolumn{2}{c}{512}       &  & \multicolumn{2}{c}{1024}      &  & \multicolumn{2}{c}{2048}        &  & \multicolumn{2}{c}{4096}                             \\ \cline{2-3} \cline{5-6} \cline{8-9} \cline{11-12} \cline{14-15} \cline{17-18} 
\multicolumn{1}{l|}{}                          & PPL           & ACC.          &  & PPL           & ACC.          &  & PPL           & ACC.          &  & PPL           & ACC.          &  & PPL             & ACC.          &  & PPL                        & ACC.                    \\ \hline
\multicolumn{1}{l|}{GAU-α   w/ I.S.}           & \textbf{2.16} & \textbf{0.84} &  & \textbf{2.48} & \textbf{0.81} &  & \textbf{2.71} & \textbf{0.78} &  & \textbf{8.60} & \textbf{0.58} &  & \textbf{109.15} & \textbf{0.23} &  & \textbf{\textgreater{}500} & \textbf{\textless{}0.1} \\
\multicolumn{1}{l|}{GAU-α   w/ softmax+}       & 2.57          & 0.80          &  & 2.85          & 0.78          &  & 3.05          & 0.76          &  & 11.02         & 0.53          &  & 145.16          & 0.19          &  & \textgreater{}500          & \textless{}0.1          \\ \bottomrule
\end{tabular}
\end{table*}
\subsubsection{InfoScale Outperforms Other Scaling Strategies for Length Extrapolation}
To further demonstrate the effectiveness of InfoScale, we compare it against Softmax Plus \cite{gau-alpha}, a similar scaling strategy. We evaluate both methods with and without CosScale on cosine attention. As shown in \cref{table 4.2.4-2} InfoScale consistently achieves higher accuracy across all conditions without CosScale. While InfoScale may exhibit slightly higher perplexity at certain sequence lengths with CosScale compared to Softmax Plus (see \cref{table 4.2.4-1}), its overall performance remains superior. This highlights the advantages of InfoScale in improving length extrapolation capabilities.

\subsection{Behaviors of CosScale}
\subsubsection{CosScale Approximates the Windowed Attention as It Increases}
\cref{f1} illustrates the impact of increasing CosScale on the attention matrix. The histograms display the distribution of $\cos\theta$, with the x-axis representing the cosine similarity. As CosScale increases, the distribution of $\cos\theta$ becomes sharper and approaches 1. At CosScale = 600, the distribution shifts left. We hypothesize that this is potentially due to numerical overflow in the gradients during the forward pass. Consequently, attention becomes increasingly concentrated around the diagonal of the attention matrix. The subsequent analyses will exclude the anomalous case of CosScale = 600.

We observe that the histogram of the query-key multiplication ($QK^T$) before applying RoPE becomes progressively sharper with larger CosScale values. This results in the attention matrix adopting a shape similar to that of the Windowed Attention, where attention is primarily focused on a local neighborhood around each token. Notably, the peak value of the attention histogram decreases at CosScale = 600, likely due to numerical precision limitations exceeding the model's capacity.
\subsubsection{Larger CosScale Favors Longer Context Windows}
\label{Larger CosScale Favors Longer Context Windows}
\cref{table 4.3.2} presents the validation performance across different CosScale values. When extrapolating to a sequence length of 4k, the optimal performance (maximum ACC and minimum PPL) generally shifts towards larger CosScale values. For example, at a sequence length of 128, CosScale = 32 yields the best performance, while at 1024, CosScale = 256 performs optimally. This trend can be attributed to the sharper angular distribution of the attention matrix with larger scales (\cref{f1}), enabling the model to focus more effectively on relevant regions within the context. However, the optimal CosScale falls back to 128 when extrapolating to 4096, possibly due to reaching the limits of the model's gradient handling capacity. Overall, a CosScale of 128 achieves the highest average accuracy across different sequence lengths.

\subsection{Comparison of the Windowed Attention and Cosine Attention with Large CosScale}
While increasing CosScale can improve length extrapolation performance, we observe that it does not reach the same level of performance as the Windowed Attention. As discussed in the previous section, this limitation may be attributed to gradient vanishing at excessively large scales. We illustrate this phenomenon with an example in \cref{f2}. The attention matrix exhibits larger values concentrated around the diagonal. However, in \cref{f2}(b) CosScale (scale = 128) displays "vertical stripe" patterns. This observation indicates that many tokens share similar attention weights for a given token, and reduces attention distinguishability within a fixed context window. Ideally, the token shading intensity should represent attention values.\cref{f2}(a) demonstrates that for the selected token highlighted in gray background, other tokens exhibit more concentrated and reasonable attention distributions. Therefore, we use a CosScale of 16 rather than 128 for the Windowed Attention method.

\cref{f2} suggests that for length extrapolation with long context windows, the dilution of attention scores caused by the introduction of new tokens is a primary factor limiting model performance. This seems to be contradictory to the findings in \cref{Larger CosScale Favors Longer Context Windows}. While increasing CosScale can help alleviate the dilution effect to some extent, it may also lead to vanishing gradient problems, which hinder further improvements. In contrast, the Windowed Attention preserves a fixed context window during training, effectively avoiding this dilution issue. In this scenario, using a smaller CosScale can enhance the gradient of the softmax function, leading to better performance.
\subsection{Ablation Study}
This section evaluates the individual contributions of InfoScale and CosScale on the GAU-α architecture. We calculate the model’s performance with: (1) neither InfoScale nor CosScale, (2) InfoScale only, (3) CosScale only, and (4) both InfoScale and CosScale. As shown in \cref{table 4.5}, the combination of InfoScale and CosScale achieves the best performance on sequences longer than the pre-training length of 512, and demonstrates the strong extrapolation capabilities of our two scale temperatures. 
\section{Conclusion and Future Work}
This paper introduces two novel scale temperatures for self-attention: InfoScale and CosScale. Our empirical results demonstrate that combining InfoScale with CosScale significantly improves the performance of eight existing length extrapolation methods. Furthermore, we show that as CosScale increases, the attention mechanism approximates the Windowed Attention. This observation suggests that attention score dilution is a critical factor limiting performance in long-range context modeling.

Future work will explore several avenues for improvement. First, we aim to enhance the capabilities of CosScale for RoPE-based language models by investigating how to promote the learning of effective relationships between tokens during training while preserving its resistance to attention score leakage during length extrapolation. This may involve a comprehensive investigation of the interplay between CosScale and the positional encoding mechanism. Second, we plan to develop a unified framework that seamlessly integrates InfoScale and CosScale for further performance gains.

While InfoScale is generally applicable to various attention mechanisms, the theoretical analysis of CosScale is specifically tailored to RoPE-based language models like GAU-α and LLaMA. This is because cosine attention focuses on angular relationships between vectors, and is naturally suited to RoPE-based models. In contrast, the vanilla self-attention mechanism in non-RoPE-based models like GPT-3 trains both angle and magnitude, potentially obscuring the full benefits of CosScale. Exploring the applicability of CosScale to these models may require further theoretical analysis and empirical investigation.

In conclusion, this work provides valuable insights into the challenges of length extrapolation in LLMs. By developing a more comprehensive theoretical understanding of self-attention, we aim to further enhance the capabilities of language models in handling long-range contexts.


\section*{Acknowledgements}

This work is supported by National Natural Science Foundation of China (No. 62072212), Development Project of Jilin Province of China (No. 20220508125RC), Guizhou Provincial Science and Technology Projects (ZK2023-297), the Science and Technology Foundation of Health Commission of Guizhou Province (gzwkj2023-565), and the Fundamental Research Funds for the Central Universities (JLU).

\section*{Impact Statement}

Our work aims to address the technical challenge of length extrapolation within the field of natural language processing. As such, the theoretical aspects and the models themselves do not directly involve ethical concerns. The results of this work are applicable to a wide range of scenarios, including human-computer interaction, financial forecasting, and genomic sequence analysis. However, if misused, they could lead to negative ethical and societal impacts, such as the dissemination of incorrect information that misleads users, unfair investment decisions resulting in adverse economic effects, and incorrect genomic sequence predictions that pose risks to health and safety. These potential impacts are not specific to this particular study but are relevant across all research within the natural language processing field.


\bibliography{main}

\begin{thebibliography}{24}
\providecommand{\natexlab}[1]{#1}
\providecommand{\url}[1]{\texttt{#1}}
\expandafter\ifx\csname urlstyle\endcsname\relax
  \providecommand{\doi}[1]{doi: #1}\else
  \providecommand{\doi}{doi: \begingroup \urlstyle{rm}\Url}\fi

\bibitem[Chen et~al.(2023{\natexlab{a}})Chen, Li, Meng, Liang, and Bing]{chen2023clex}
Chen, G., Li, X., Meng, Z., Liang, S., and Bing, L.
\newblock Clex: Continuous length extrapolation for large language models.
\newblock \emph{arXiv preprint arXiv:2310.16450}, 2023{\natexlab{a}}.

\bibitem[Chen et~al.(2023{\natexlab{b}})Chen, Wong, Chen, and Tian]{chen2023extending}
Chen, S., Wong, S., Chen, L., and Tian, Y.
\newblock Extending context window of large language models via positional interpolation.
\newblock \emph{arXiv preprint arXiv:2306.15595}, 2023{\natexlab{b}}.

\bibitem[Chiang \& Cholak(2022)Chiang and Cholak]{chiang2022overcoming}
Chiang, D. and Cholak, P.
\newblock Overcoming a theoretical limitation of self-attention.
\newblock \emph{arXiv preprint arXiv:2202.12172}, 2022.

\bibitem[Deng et~al.(2019)Deng, Guo, Xue, and Zafeiriou]{deng2019arcface}
Deng, J., Guo, J., Xue, N., and Zafeiriou, S.
\newblock Arcface: Additive angular margin loss for deep face recognition.
\newblock In \emph{Proceedings of the IEEE/CVF conference on computer vision and pattern recognition}, pp.\  4690--4699, 2019.

\bibitem[Glorot \& Bengio(2010)Glorot and Bengio]{glorot2010understanding}
Glorot, X. and Bengio, Y.
\newblock Understanding the difficulty of training deep feedforward neural networks.
\newblock In \emph{Proceedings of the thirteenth international conference on artificial intelligence and statistics}, pp.\  249--256. JMLR Workshop and Conference Proceedings, 2010.

\bibitem[Han et~al.(2024)Han, Wang, Peng, Xiong, Chen, Ji, and Wang]{han2024lm}
Han, C., Wang, Q., Peng, H., Xiong, W., Chen, Y., Ji, H., and Wang, S.
\newblock Lm-infinite: Zero-shot extreme length generalization for large language models.
\newblock In \emph{Proceedings of the 2024 Conference of the North American Chapter of the Association for Computational Linguistics: Human Language Technologies (Volume 1: Long Papers)}, pp.\  3991--4008, 2024.

\bibitem[Kolmogorov \& Bharucha-Reid(2018)Kolmogorov and Bharucha-Reid]{kolmogorov2018foundations}
Kolmogorov, A.~N. and Bharucha-Reid, A.~T.
\newblock \emph{Foundations of the theory of probability: Second English Edition}.
\newblock Courier Dover Publications, 2018.

\bibitem[Kruschke(2010)]{kruschke2010bayesian}
Kruschke, J.~K.
\newblock Bayesian data analysis.
\newblock \emph{Wiley Interdisciplinary Reviews: Cognitive Science}, 1\penalty0 (5):\penalty0 658--676, 2010.

\bibitem[Peng et~al.(2023)Peng, Quesnelle, Fan, and Shippole]{peng2023yarn}
Peng, B., Quesnelle, J., Fan, H., and Shippole, E.
\newblock Yarn: Efficient context window extension of large language models.
\newblock \emph{arXiv preprint arXiv:2309.00071}, 2023.

\bibitem[Press et~al.(2021)Press, Smith, and Lewis]{press2021train}
Press, O., Smith, N.~A., and Lewis, M.
\newblock Train short, test long: Attention with linear biases enables input length extrapolation.
\newblock \emph{arXiv preprint arXiv:2108.12409}, 2021.

\bibitem[Shannon(1948)]{shannon1948mathematical}
Shannon, C.~E.
\newblock A mathematical theory of communication.
\newblock \emph{The Bell system technical journal}, 27\penalty0 (3):\penalty0 379--423, 1948.

\bibitem[Stewart(2012)]{stewart2012calculus}
Stewart, J.
\newblock \emph{Calculus: early transcendentals}.
\newblock Cengage Learning, 2012.

\bibitem[Su(2021)]{kexuefm-8823}
Su, J.
\newblock Analyzing the scale operation of attention from the perspective of entropy invariance(in chinese).
\newblock Technical report, Dec 2021.
\newblock URL \url{https://kexue.fm/archives/8823}.

\bibitem[Su(2023{\natexlab{a}})]{kexuefm-9812}
Su, J.
\newblock Scale operation of attention from the perspective of gradient maximization(in chinese).
\newblock Technical report, Oct 2023{\natexlab{a}}.
\newblock URL \url{https://spaces.ac.cn/archives/9812}.

\bibitem[Su(2023{\natexlab{b}})]{rerope2023}
Su, J.
\newblock Rectified rotary position embeddings.
\newblock \url{https://github.com/bojone/rerope}, 2023{\natexlab{b}}.

\bibitem[Su et~al.(2022)Su, Pan, Wen, and Liu]{gau-alpha}
Su, J., Pan, S., Wen, B., and Liu, Y.
\newblock Gau-α: Gau-based transformers for nlp - zhuiyiai.
\newblock Technical report, 2022.
\newblock URL \url{https://github.com/ZhuiyiTechnology/GAU-alpha}.

\bibitem[Touvron et~al.(2023)Touvron, Lavril, Izacard, Martinet, Lachaux, Lacroix, Rozi{\`e}re, Goyal, Hambro, Azhar, et~al.]{touvron2023llama}
Touvron, H., Lavril, T., Izacard, G., Martinet, X., Lachaux, M.-A., Lacroix, T., Rozi{\`e}re, B., Goyal, N., Hambro, E., Azhar, F., et~al.
\newblock Llama: Open and efficient foundation language models.
\newblock \emph{arXiv preprint arXiv:2302.13971}, 2023.

\bibitem[Vaswani(2017)]{vaswani2017attention}
Vaswani, A.
\newblock Attention is all you need.
\newblock \emph{Advances in Neural Information Processing Systems}, 2017.

\bibitem[Wang et~al.(2018{\natexlab{a}})Wang, Cheng, Liu, and Liu]{wang2018additive}
Wang, F., Cheng, J., Liu, W., and Liu, H.
\newblock Additive margin softmax for face verification.
\newblock \emph{IEEE Signal Processing Letters}, 25\penalty0 (7):\penalty0 926--930, 2018{\natexlab{a}}.

\bibitem[Wang et~al.(2018{\natexlab{b}})Wang, Wang, Zhou, Ji, Gong, Zhou, Li, and Liu]{wang2018cosface}
Wang, H., Wang, Y., Zhou, Z., Ji, X., Gong, D., Zhou, J., Li, Z., and Liu, W.
\newblock Cosface: Large margin cosine loss for deep face recognition.
\newblock In \emph{Proceedings of the IEEE conference on computer vision and pattern recognition}, pp.\  5265--5274, 2018{\natexlab{b}}.

\bibitem[Wang et~al.(2024)Wang, Ma, Dong, Huang, Zhang, and Wei]{wang2024deepnet}
Wang, H., Ma, S., Dong, L., Huang, S., Zhang, D., and Wei, F.
\newblock Deepnet: Scaling transformers to 1,000 layers.
\newblock \emph{IEEE Transactions on Pattern Analysis and Machine Intelligence}, 2024.

\bibitem[Xiao et~al.(2023)Xiao, Tian, Chen, Han, and Lewis]{xiao2023efficient}
Xiao, G., Tian, Y., Chen, B., Han, S., and Lewis, M.
\newblock Efficient streaming language models with attention sinks.
\newblock \emph{arXiv preprint arXiv:2309.17453}, 2023.

\bibitem[Yu(2023)]{yu2023paraphrasing}
Yu, Y.
\newblock " paraphrasing the original text" makes high accuracy long-context qa.
\newblock \emph{arXiv preprint arXiv:2312.11193}, 2023.

\bibitem[Zhu et~al.(2023)Zhu, Yang, Wang, Song, Wu, Wei, and Li]{zhu2023pose}
Zhu, D., Yang, N., Wang, L., Song, Y., Wu, W., Wei, F., and Li, S.
\newblock Pose: Efficient context window extension of llms via positional skip-wise training.
\newblock \emph{arXiv preprint arXiv:2309.10400}, 2023.

\end{thebibliography}
\bibliographystyle{icml2025}

\newpage
\appendix
\onecolumn
\section{Appendix}
\label{appendix}
\subsection{Dataset}
\label{Dataset}
\subsubsection{Dataset Partitioning}
We randomly select 80\% of the samples in the WJ dataset as the training set, and employ a stratified sampling strategy to extract 10 out of the remaining samples as the test set. All the other samples are designated as the validation set. The stratified sampling strategy is described the next section and \cref{table 9.1.1} provides details of the partitioned sets.
\begin{table*}[h]
\caption{Details of the WJ dataset.}
\label{table 9.1.1}
\centering
\vskip 0.1in
\begin{tabular}{cccc}
\toprule
\multicolumn{1}{l}{} & Train  & Evaluation & Test  \\
\midrule
number of   samples  & 13372  & 3333       & 10    \\
average length       & 20121  & 20146      & 18971 \\
min length           & 10815  & 13971      & 20987 \\
max length           & 205672 & 161019     & 15959\\
\bottomrule
\end{tabular}
\end{table*}
\subsubsection{Stratified 10 Samples as Test Set}
\begin{figure}[h]
\vskip 0.1in
\begin{center}
\centerline{\includegraphics[width=\textwidth]{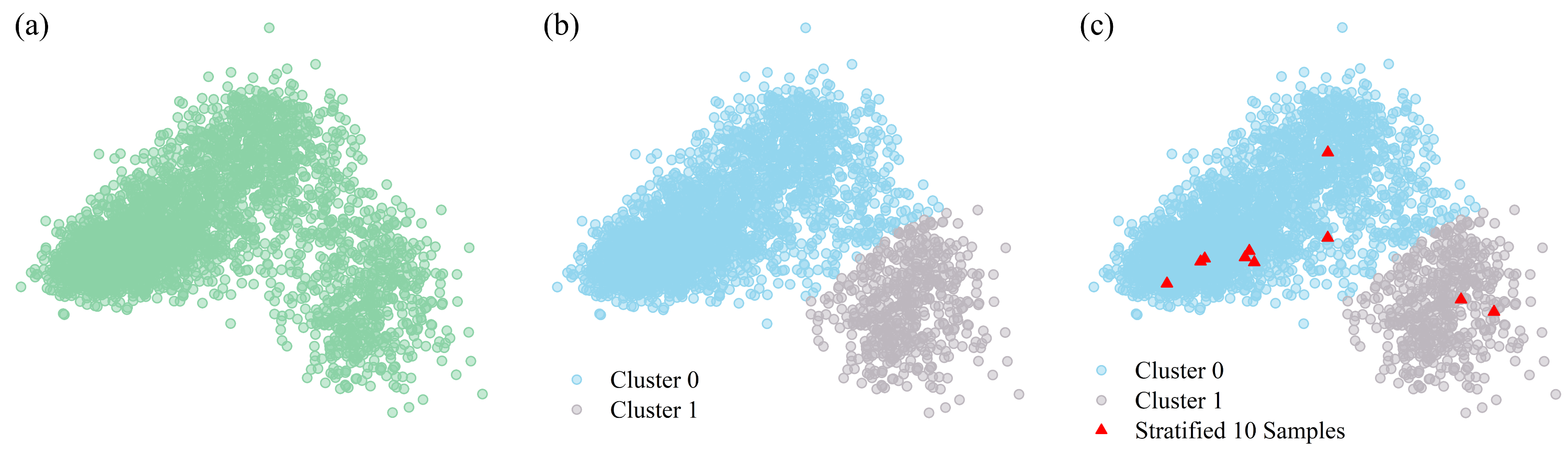}}
\caption{Visualization of spectral clustering with stratified 10 samples as test set. (a) Distribution of 2D PCA features for different samples. (b) Clustering results. (c) Sample distribution obtained through stratified random sampling.}
\label{f3}
\end{center}
\vskip -0.1in
\end{figure}

First, we apply Principal Component Analysis (PCA) to reduce the dimensionality of each sample's feature representation from 768 to 2. This allows for visualization and identification of clusters within the data. Based on the observed clustering tendency, we utilize spectral clustering to partition the dataset into two distinct clusters. Within each cluster, we perform stratified sampling to select a total of 10 representative samples across both clusters (\cref{f3}). These 10 samples constitute a test set, denoted as WJ-Unique10, which is used for independent testing.
\subsection{Model Training and Fine-tuning Process}
\label{Model Training and Fine-tuning Process}
\subsubsection{Training Process}
\begin{figure}[]
\vskip 0.1in
\begin{center}
\centerline{\includegraphics[width=0.4\textwidth]{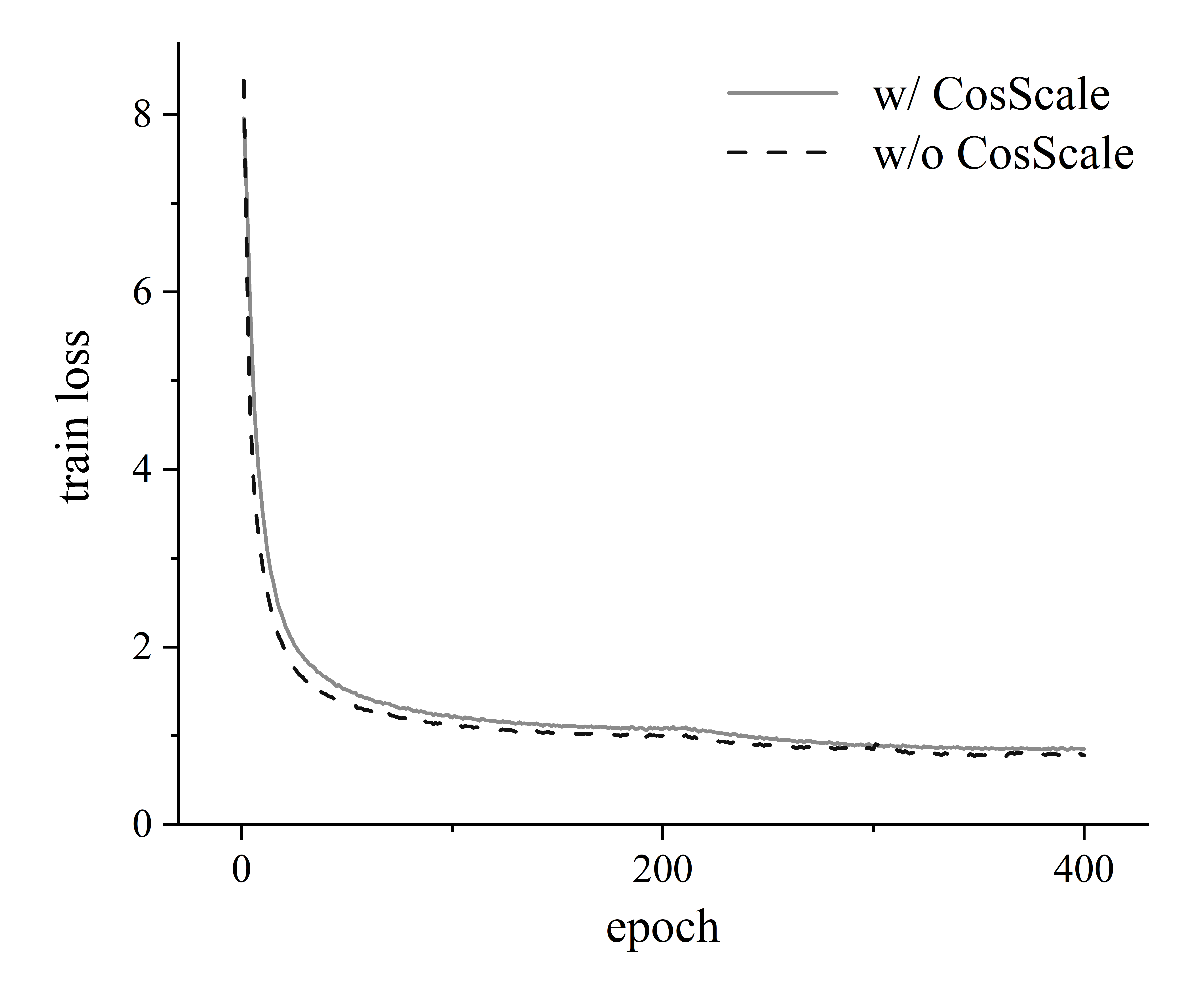}}
\caption{Training loss of the base GAU-α model with and without CosScale=128. The x-axis represents the epoch.}
\label{f4}
\end{center}
\vskip -0.1in
\end{figure}

We train the model with 400 epochs and save the best checkpoint according to the lowest evaluation loss. \cref{f4} illustrates the training process of the base GAU-α model, and shows the training loss at the end of each epoch. The solid line displays the results for the GAU-α model with CosScale = 128, while the dashed line shows the results without CosScale.
It turns out that the training loss of CosScale descending slightly slower than the vanilla one. We speculate that CosScale = 128 is a much large value that hinder the speed of training process.
\subsubsection{Fine-tuning Process}
\label{Fine-tuning Process}
\begin{figure}[]
\vskip 0.1in
\begin{center}
\centerline{\includegraphics[width=0.8\textwidth]{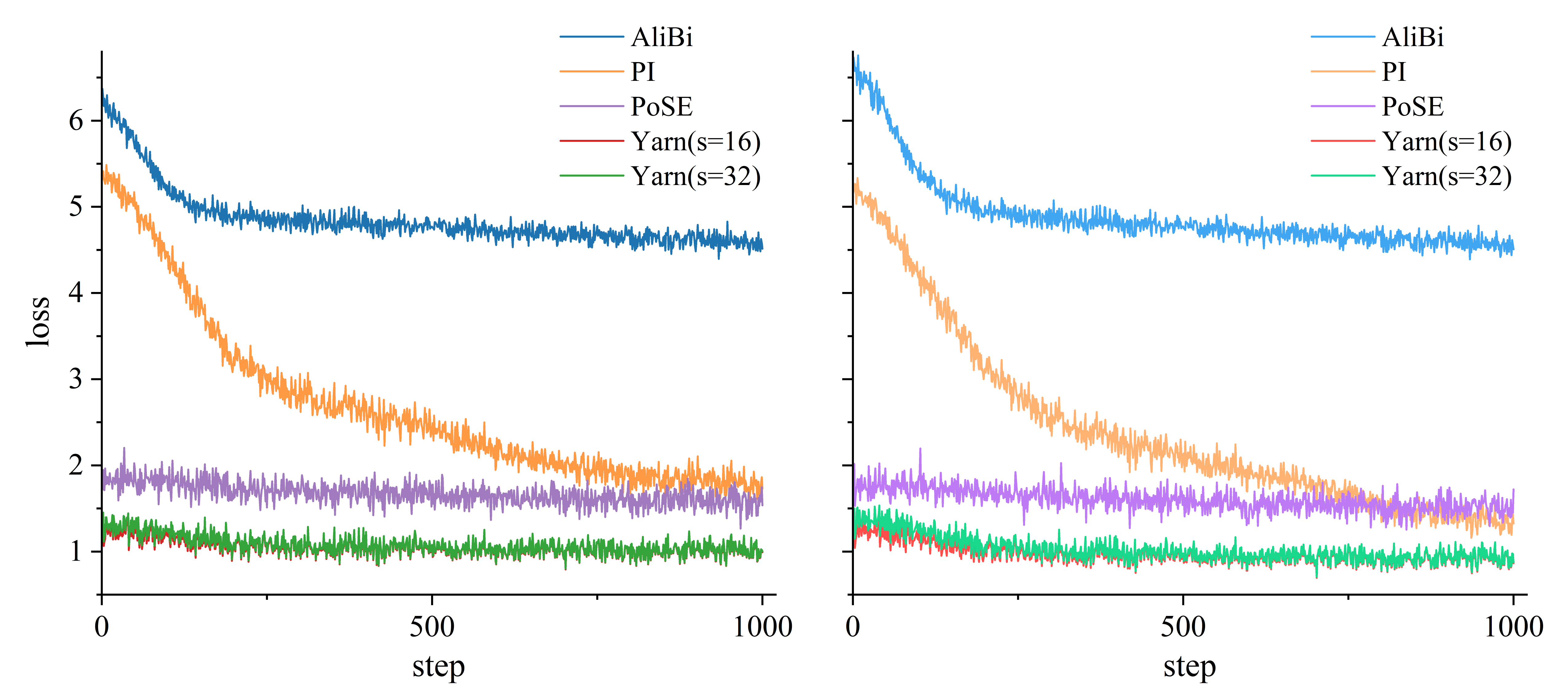}}
\caption{Training loss of all the baseline models integrating InfoScale. The left panel shows the baselines with CosScale, while the right panel shows the baselines without CosScale. All baseline models requiring fine-tuning were fine-tuned for 1000 steps (equal to 100 epochs). The x-axis represents the training steps.}
\label{f5}
\end{center}
\vskip -0.1in
\end{figure}

We compare the baseline models integrating InfoScale with several baselines on the length extrapolation task, all trained using the WJ dataset. \cref{f5} presents the training loss for all models that require fine-tuning. For methods requiring training or fine-tuning, we followed these procedures:
\begin{itemize}

   \item \textbf{{PI:}} Following \cite{chen2023extending}, we trained PI with a maximum sequence length of 64 and a scaling factor of t=4.
  \item \textbf{{YaRN:}} The model was trained with scaling factors of 16 and 32, as per \cite{peng2023yarn}.
   \item \textbf{{ALiBi:}} We removed RoPE from GAU-α and trained a model with an attention bias.
  \item \textbf{{PoSE:}} We used a preset extrapolation length of 4096.

\end{itemize}
For methods not requiring training or fine-tuning:
\begin{itemize}

\item\textbf{{ReRoPE:}} We replaced the original RoPE in GAU-α with a ReRoPE implementation. Note that the RoPE implementation in LLaMA differs from that in GAU-α.

\item\textbf{{StreamingLLM:}} Following \cite{xiao2023efficient}, we set the initial 4 tokens as attention sink tokens and used a sliding window approach to anchor attention computation.

\item\textbf{{LM-Infinite:}} The number of reserved initial tokens was set to 5, based on the recommendations in \cite{han2024lm}.

\item\textbf{{Windowed Attention:}} We set the window size to match the training sequence length (64) and masked the attention matrix outside the window with -INF.
\end{itemize}
\subsection{Evaluation Method}
\label{Evaluation Method}
We evaluate the performance of our proposed methods using two standard metrics commonly employed in the length extrapolation task:

\textbf{Accuracy (ACC):} To assess the model's precision in predicting the correct token at each position, we calculate the token-level accuracy. This metric provides a direct measure of the model's ability to generate accurate sequences.

\textbf{Perplexity (PPL): }Perplexity quantifies the uncertainty of a language model's predictions. It is calculated as the exponential of the cross-entropy loss. Lower perplexity values indicate better performance, as they suggest that the model assigns higher probabilities to the correct tokens in a sequence.
\subsection{Detailed Derivations of InfoScale}
\label{The details of the deduction of InfoScale}
\textbf{\cref{ass1}} We assume that the attention scores follow the Law of Large Numbers \cite{kolmogorov2018foundations}. This implies that the average of a large number of attention scores converges to the expected value of the attention score distribution. Formally, this is expressed as: ${\sum_{i = 1}^{n}y_{i}} = n \times \frac{1}{n}{\sum_{i = 1}^{n}y_{i}} = n\mathbb{E}\left\lbrack y_{i} \right\rbrack$

Applying \cref{ass1} to Eq. \ref{eq5}, we obtain the following equation:
\begin{align*}
H_{i} &= - {\sum\limits_{j = 1}^{n}{\beta_{ij}{\log\left( \beta_{ij} \right)}}}\\
&= - {\sum\limits_{j = 1}^{n}{\frac{e^{\lambda\boldsymbol{q}_{{i}}\boldsymbol{k}_{{j}}^{{T}}}}{\sum\limits_{j = 1}^{n}e^{\lambda \boldsymbol{q}_{i}\boldsymbol{k}_{j}^{T}}}{\log\left( \frac{e^{\lambda \boldsymbol{q}_{i}\boldsymbol{k}_{j}^{T}}}{\sum\limits_{j = 1}^{n}e^{\lambda \boldsymbol{q}_{i}\boldsymbol{k}_{j}^{T}}} \right)}}}\\
&= - {\sum\limits_{j = 1}^{n}{\frac{e^{\lambda \boldsymbol{q}_{i}\boldsymbol{k}_{j}^{T}}}{\sum\limits_{j = 1}^{n}e^{\lambda \boldsymbol{q}_{i}\boldsymbol{k}_{j}^{T}}}\left( \lambda \boldsymbol{q}_{i}\boldsymbol{k}_{j}^{T} - {\log\left( {\sum\limits_{j = 1}^{n}e^{\lambda \boldsymbol{q}_{i}\boldsymbol{k}_{j}^{T}}} \right)} \right)}}\\
&= - {\sum\limits_{j = 1}^{n}\frac{\lambda \boldsymbol{q}_{i}\boldsymbol{k}_{j}^{T}e^{\lambda \boldsymbol{q}_{i}\boldsymbol{k}_{j}^{T}}}{\sum\limits_{j = 1}^{n}e^{\lambda \boldsymbol{q}_{i}\boldsymbol{k}_{j}^{T}}}} + {\sum\limits_{j = 1}^{n}\frac{e^{\lambda \boldsymbol{q}_{i}\boldsymbol{k}_{j}^{T}}}{\sum\limits_{j = 1}^{n}e^{\lambda \boldsymbol{q}_{i}\boldsymbol{k}_{j}^{T}}}}{\log\left( {\sum\limits_{j = 1}^{n}e^{\lambda \boldsymbol{q}_{i}\boldsymbol{k}_{j}^{T}}} \right)}\\
&= - \frac{\sum\limits_{j = 1}^{n}{\lambda\boldsymbol{q}_{i}\boldsymbol{k}_{j}^{T}e^{\lambda\boldsymbol{q}_{i}\boldsymbol{k}_{j}^{T}}}}{\sum\limits_{j = 1}^{n}e^{\lambda\boldsymbol{q}_{i}\boldsymbol{k}_{j}^{T}}} + {\log\left( {\sum\limits_{j = 1}^{n}e^{\lambda\boldsymbol{q}_{i}\boldsymbol{k}_{j}^{T}}} \right)}
\end{align*}
Here we use sampling for estimation:
\begin{gather*}
{\sum\limits_{j = 1}^{n}{\lambda\boldsymbol{q}_{i}\boldsymbol{k}_{j}^{T}e^{\lambda\boldsymbol{q}_{i}\boldsymbol{k}_{j}^{T}}}} = n \times \frac{1}{n} \times {\sum\limits_{j = 1}^{n}{\lambda\boldsymbol{q}_{i}\boldsymbol{k}_{j}^{T}e^{\lambda\boldsymbol{q}_{i}\boldsymbol{k}_{j}^{T}}}} = n\mathbb{E}\left\lbrack \lambda\boldsymbol{q}_{i}\boldsymbol{k}_{j}^{T}e^{\lambda\boldsymbol{q}_{i}\boldsymbol{k}_{j}^{T}} \right\rbrack\\
Z = {\sum\limits_{j = 1}^{n}e^{\lambda\boldsymbol{q}_{i}\boldsymbol{k}_{j}^{T}}} = n \times \frac{1}{n} \times {\sum\limits_{j = 1}^{n}e^{\lambda\boldsymbol{q}_{i}\boldsymbol{k}_{j}^{T}}} = n\mathbb{E}\left\lbrack e^{\lambda\boldsymbol{q}_{i}\boldsymbol{k}_{j}^{T}} \right\rbrack
\end{gather*}
Then we have:
\begin{align*}
    H_{i} &\cong - \frac{\sum\limits_{j = 1}^{n}{\lambda\boldsymbol{q}_{i}\boldsymbol{k}_{j}^{T}e^{\lambda\boldsymbol{q}_{i}\boldsymbol{k}_{j}^{T}}}}{\sum\limits_{j = 1}^{n}e^{\lambda\boldsymbol{q}_{i}\boldsymbol{k}_{j}^{T}}} + {\log\left( {\sum\limits_{j = 1}^{n}e^{\lambda\boldsymbol{q}_{i}\boldsymbol{k}_{j}^{T}}} \right)}\\
    &= {\log\left( {n\mathbb{E}\left\lbrack e^{\lambda\boldsymbol{q}_{i}\boldsymbol{k}_{j}^{T}} \right\rbrack} \right)} - \frac{n\mathbb{E}\left\lbrack {\lambda\boldsymbol{q}_{i}\boldsymbol{k}_{j}^{T}e^{\lambda\boldsymbol{q}_{i}\boldsymbol{k}_{j}^{T}}} \right\rbrack}{Z}\\
    &= {\log(n)} + {\log\left( {\mathbb{E}\left\lbrack e^{\lambda\boldsymbol{q}_{i}\boldsymbol{k}_{j}^{T}} \right\rbrack} \right)} - \frac{n\mathbb{E}\left\lbrack {\lambda\boldsymbol{q}_{i}\boldsymbol{k}_{j}^{T}e^{\lambda\boldsymbol{q}_{i}\boldsymbol{k}_{j}^{T}}} \right\rbrack}{n\mathbb{E}\left\lbrack e^{\lambda\boldsymbol{q}_{i}\boldsymbol{k}_{j}^{T}} \right\rbrack}\\
    H_{i} &\cong \widehat{H_{i}} = {\log(n)} + {\log\left( {\mathbb{E}\left\lbrack e^{\lambda\boldsymbol{q}_{i}\boldsymbol{k}_{j}^{T}} \right\rbrack} \right)} - \frac{\mathbb{E}\left\lbrack {\lambda\boldsymbol{q}_{i}\boldsymbol{k}_{j}^{T}e^{\lambda\boldsymbol{q}_{i}\boldsymbol{k}_{j}^{T}}} \right\rbrack}{\mathbb{E}\left\lbrack e^{\lambda\boldsymbol{q}_{i}\boldsymbol{k}_{j}^{T}} \right\rbrack}
\end{align*}
where $\widehat{H_{i}}$ is the approximation of $\widehat{H_{i}}$ by the Law of Large Numbers \cite{kolmogorov2018foundations}. Now we have two expectations $\mathbb{E}\left\lbrack e^{\lambda\boldsymbol{q}_{i}\boldsymbol{k}_{j}^{T}} \right\rbrack$ and $\mathbb{E}\left\lbrack {\lambda\boldsymbol{q}_{i}\boldsymbol{k}_{j}^{T}e^{\lambda\boldsymbol{q}_{i}\boldsymbol{k}_{j}^{T}}} \right\rbrack$ to solve, but they are not directly computable. Fortunately, according to the weight initialization strategy \cite{glorot2010understanding}, we can reasonably apply \cref{ass2} to make further derivations.

\textbf{\cref{ass2}} The word embedding is laid in a $d_{k}$-dimensional $r = \sqrt{vd_{k}}$ rational hypersphere, and then we get the expectation of $\boldsymbol{q}_{i}\boldsymbol{k}_{j}^{T}$ corresponds to $r^{2}{\cos\theta}$, where $\theta$ is the angle between $q_{i}$ and $k_{j}$. Here $v$ is the variance of the word embedding for weight initialization. 

Let $\lambda \boldsymbol{q}_{i}\boldsymbol{k}_{j} = \lambda r^{2}{\cos\theta} = \lambda vd_{k}{\cos\theta} = \alpha{\cos\theta}$, then we have:
\begin{align*}
    \widehat{H_{i}} &= {\log(n)} + {\log\left( {\mathbb{E}\left\lbrack e^{\lambda\boldsymbol{q}_{i}\boldsymbol{k}_{j}^{T}} \right\rbrack} \right)} - \frac{\mathbb{E}\left\lbrack {\lambda\boldsymbol{q}_{i}\boldsymbol{k}_{j}^{T}e^{\lambda\boldsymbol{q}_{i}\boldsymbol{k}_{j}^{T}}} \right\rbrack}{\mathbb{E}\left\lbrack e^{\lambda\boldsymbol{q}_{i}\boldsymbol{k}_{j}^{T}} \right\rbrack}\\
    &= {\log(n)} + {\log\left( \mathbb{E}\left\lbrack e^{\alpha{\cos\theta}} \right\rbrack \right)} - \frac{\mathbb{E}\left\lbrack {\alpha{\cos\theta}e^{\alpha{\cos\theta}}} \right\rbrack}{\mathbb{E}\left\lbrack e^{\alpha{\cos\theta}} \right\rbrack}
\end{align*}
Consider the probability density function of a \textit{n}-dimension hypersphere:
\begin{align}
    p_{n}(\theta) = \frac{\Gamma\left( \frac{n}{2} \right)}{\Gamma\left( \frac{n - 1}{2} \right)\sqrt{\pi}}{\sin^{n - 2}(\theta)}
\end{align}
where $\Gamma(n) = \left( {n - 1} \right)!$. Then we get
\begin{align*}
    H_{i} \cong \widehat{H_{i}} &= {\log(n)} + {\log\left( {\mathbb{E}\left\lbrack e^{\alpha{\cos\theta}} \right\rbrack} \right)} - \frac{\mathbb{E}\left\lbrack {\alpha{\cos\theta}e^{\alpha{\cos\theta}}} \right\rbrack}{\mathbb{E}\left\lbrack e^{\alpha{\cos\theta}} \right\rbrack}\mathrm{d}\theta\\
    &= {\log(n)} + {\log\left( \frac{\int_{0}^{\pi}{e^{\alpha{\cos\theta}}p_{d_{k}}(\theta)\mathrm{d}\theta}}{\int_{0}^{\pi}{p_{d_{k}}(\theta)\mathrm{d}\theta}} \right)} - \frac{\tfrac{\int_{0}^{\pi}{\alpha{\cos\theta}e^{\alpha{\cos\theta}}p_{d_{k}}(\theta)\mathrm{d}\theta}}{\int_{0}^{\pi}{p_{d_{k}}(\theta)\mathrm{d}\theta}}}{\tfrac{\int_{0}^{\pi}{e^{\alpha{\cos\theta}}p_{d_{k}}(\theta)\mathrm{d}\theta}}{\int_{0}^{\pi}{p_{d_{k}}(\theta)\mathrm{d}\theta}}}\\
    &= {\log(n)} + {\log\left( \frac{\int_{0}^{\pi}{e^{\alpha{\cos\theta}}p_{d_{k}}(\theta)\mathrm{d}\theta}}{\int_{0}^{\pi}{p_{d_{k}}(\theta)\mathrm{d}\theta}} \right)} - \frac{\int_{0}^{\pi}{\alpha{\cos\theta}e^{\alpha{\cos\theta}}p_{d_{k}}(\theta)\mathrm{d}\theta}}{\int_{0}^{\pi}{e^{\alpha{\cos\theta}}p_{d_{k}}(\theta)\mathrm{d}\theta}}\\
    &= {\log(n)} + {\log\left( \frac{\int_{0}^{\pi}{e^{\alpha{\cos\theta}}{\sin^{d_{k} - 2}\theta}\mathrm{d}\theta}}{\int_{0}^{\pi}{{\sin^{d_{k} - 2}\theta}\mathrm{d}\theta}} \right)} - \frac{\int_{0}^{\pi}{\alpha{\cos\theta}e^{\alpha{\cos\theta}}{\sin^{d_{k} - 2}\theta}\mathrm{d}\theta}}{\int_{0}^{\pi}{e^{\alpha{\cos\theta}}{\sin^{d_{k} - 2}\theta}\mathrm{d}\theta}}
\end{align*}
There are three integrals. We introduce \cref{ass3} to solve these integrals:

\textbf{\cref{ass3}} The integrals can be approximated by Laplace approximation \cite{kruschke2010bayesian}.
Finally, we get:
\begin{align*}
    {\int_{0}^{\pi}{e^{\alpha{\cos\theta}}{\sin^{d_{k} - 2}\theta}\mathrm{d}\theta}} = {\int_{0}^{\pi}{e^{\alpha{\cos\theta}}e^{\log{({\sin^{d_{k} - 2}\theta})}}\mathrm{d}\theta}} = {\int_{0}^{\pi}{e^{\alpha{\cos\theta} + (d_{k} - 2)log({\sin\theta})}\mathrm{d}\theta}}
\end{align*}
When $\left. d_{k}\rightarrow\infty \right.$, $d_{k} \approx d_{k} - 2$, then we get
\begin{align*}
    g_{1}(\theta) = \alpha{\cos\theta} + \left( {d_{k} - 2} \right){\log\left( {\sin\theta} \right)} \approx \alpha{\cos\theta} + d_{k}{\log\left( {\sin\theta} \right)}
\end{align*}
Then we have
\begin{gather*}
    g_{1}(\theta)^{'} = - \alpha{\cos\theta} + \frac{d_{k}{\cos\theta}}{\sin\theta} = 0\\
    {\cos\theta_{1}^{*}} = \frac{\lambda r^{2}}{d}\left( {1 - {\cos^{2}\theta_{1}^{*}}} \right) = \frac{\sqrt{4\lambda^{2}r^{4} + d_{k}^{2}} - d_{k}}{2\lambda r^{2}}
\end{gather*}
When $r^{2} = vd$:
\begin{gather*}
    {\cos\theta_{1}^{*}} = \lambda v\left( {1 - {\cos^{2}\theta_{1}^{*}}} \right) = \frac{\sqrt{4\lambda^{2}v^{2} + 1} - 1}{2\lambda v}\\
    g_{1}(\theta) = \alpha{\cos\theta_{1}^{*}} + d_k\log\left( {\sin\theta_{1}^{*}} \right) + \frac{1}{2}\left( {- \frac{\alpha}{\cos\theta_{1}^{*}} - \alpha{\cos\theta_{1}^{*}}} \right)\left( {\theta - \theta_{1}^{*}} \right)^{2}
\end{gather*}
The same as above, we have:
\begin{gather*}
    {\int_{0}^{\pi}{{\sin^{d_{k} - 2}\theta}\mathrm{d}\theta}} = {\int_{0}^{\pi}{e^{{({d_{k} - 2})}\log({\sin\theta})}\mathrm{d}\theta}}\\
    g_{2}(\theta) = \left( {d_{k} - 2} \right){\log\left( {\sin\theta} \right)} \approx d_{k}{\log\left( {\sin\theta} \right)}
\end{gather*}
Then we calculate:
\begin{align*}
    g_{2}(\theta)^{'} &= \frac{d_{k}{\cos\theta}}{\sin\theta} = 0,~\theta_{2}^{*} = \frac{\pi}{2}\\
    g_{2}(\theta) &= - \frac{d_{k}}{2}\left( {\theta - \frac{\pi}{2}} \right)^{2}
\end{align*}
Finally, we have
\begin{align*}
    {\int_{0}^{\pi}{e^{\alpha{\cos\theta}}{\sin^{d_{k} - 2}\theta}\mathrm{d}\theta}} &= {\int_{0}^{\pi}{e^{ \alpha{\cos\theta_{1}^{*}} + d_{k}{\log{({\sin{\theta_{1}^{*})}}}} + \frac{1}{2}{({- \tfrac{\alpha}{\cos\theta_{1}^{*}} - \alpha{\cos\theta_{1}^{*}}})}{({\theta - \theta_{1}^{*}})}^{2} }\mathrm{d}\theta}}\\
    &\cong {\int_{- \infty}^{\infty}{e^{\alpha{\cos\theta_{1}^{*}} + d_{k}{\log{({\sin{\theta_{1}^{*})}}}} + \frac{1}{2}{({- \tfrac{\alpha}{\cos\theta_{1}^{*}} - \alpha{\cos\theta_{1}^{*}}})}{({\theta - \theta_{1}^{*}})}^{2}}\mathrm{d}\theta}}\\
    &= e^{\alpha{\cos\theta_{1}^{*}} + d_{k}{\log{({\sin{\theta_{1}^{*})}}}}}{\int_{- \infty}^{\infty}{e^{\frac{1}{2}{({- \tfrac{\alpha}{\cos\theta_{1}^{*}} - \alpha{\cos\theta_{1}^{*}}})}{({\theta - \theta_{1}^{*}})}^{2}}\mathrm{d}\theta}}\\
    &= \sqrt{\frac{2\pi}{\alpha\left( {\frac{1}{\cos\theta_{1}^{*}} + {\cos\theta_{1}^{*}}} \right)}}e^{\alpha{\cos\theta_{1}^{*}} + d_{k}{\log{({\sin{\theta_{1}^{*})}}}}}
\end{align*}
And ${\int_{0}^{\pi}{{\sin^{d_{k} - 2}\theta}\mathrm{d}\theta}} = \sqrt{\frac{2\pi}{d_{k}}}$

Then we turn back to information entropy:
\begin{align*}
    H_{i} \cong \widehat{H_{i}} &\cong {\log(n)} + {\log\left( \frac{\int_{0}^{\pi}{e^{\alpha{\cos\theta}}{\sin^{d_{k} - 2}\theta}\mathrm{d}\theta}}{\int_{0}^{\pi}{{\sin^{d_{k} - 2}\theta}\mathrm{d}\theta}} \right)} - \frac{\int_{0}^{\pi}{\alpha{\cos\theta}e^{\alpha{\cos\theta}}{\sin^{d_{k} - 2}\theta}\mathrm{d}\theta}}{\int_{0}^{\pi}{e^{\alpha{\cos\theta}}{\sin^{d_{k} - 2}\theta}\mathrm{d}\theta}}\\
    &= {\log(n)} + {\log\left( \frac{\sqrt{\frac{2\pi}{\alpha\left( {\frac{1}{\cos\theta_{1}^{*}} + {\cos\theta_{1}^{*}}} \right)}}e^{\alpha{\cos\theta_{1}^{*}} + d_{k}{\log{({\sin{\theta_{1}^{*})}}}}}}{\sqrt{\frac{2\pi}{d_{k}}}} \right)} - \frac{\alpha{\cos\theta_{1}^{*}}g_{1}\left( \theta_{1}^{*} \right)}{g_{1}\left( \theta_{1}^{*} \right)}\\
    &= {\log(n)} + \alpha{\cos\theta_{1}^{*}} + d_{k}{\log\left( {\sin\left. \theta_{1}^{*} \right)} \right.} + {\log\left( \sqrt{\frac{d_{k}}{\alpha\left( {\frac{1}{\cos\theta_{1}^{*}} + {\cos\theta_{1}^{*}}} \right)}} \right)} - \alpha{\cos\theta_{1}^{*}}
\end{align*}
When $r^{2} = vd_{k}$:
\begin{align*}
    H_{i} \cong \widehat{H_{i}} &\cong {\log(n)} + d_{k}{\log\left( {\sin\left. \theta_{1}^{*} \right)} \right.} + {\log\left( \sqrt{\frac{1}{\lambda\left( {\frac{1}{\cos\theta_{1}^{*}} + {\cos\theta_{1}^{*}}} \right)}} \right)}\\
    &= {\log(n)} + \frac{d_{k}}{2}\left\lbrack {{\log\left( {\sqrt{4\lambda^{2}v^{2} + 1} - 1} \right)} - {\log\left( 2\lambda^{2}v^{2} \right)}} \right\rbrack - \frac{1}{2}{\log\left( {\frac{\sqrt{4\lambda^{2}v^{2} + 1} - 1}{2} + \frac{2\lambda^{2}v^{2}}{\sqrt{4\lambda^{2}v^{2} + 1} - 1}} \right)}\\
    &= {\log(n)} + \frac{d_{k}}{2}\left\lbrack {{\log\left( {\sqrt{4\lambda^{2}v^{2} + 1} - 1} \right)} - {\log\left( 2\lambda^{2}v^{2} \right)}} \right\rbrack - \frac{1}{2}{\log\left( \frac{4\lambda^{2}v^{2} + 1 + 1 - 2\sqrt{4\lambda^{2}v^{2} + 1} + 4\lambda^{2}v^{2}}{2\left( {\sqrt{4\lambda^{2}v^{2} + 1} - 1} \right)} \right)}\\
    &= {\log(n)} + \frac{d_{k}}{2}\left\lbrack {{\log\left( {\sqrt{4\lambda^{2}v^{2} + 1} - 1} \right)} - {\log\left( 2\lambda^{2}v^{2} \right)}} \right\rbrack - \frac{1}{2}{\log\left( \frac{2\sqrt{4\lambda^{2}v^{2} + 1}\left( \sqrt{4\lambda^{2}v^{2} + 1} - 1 \right)}{2\left( {\sqrt{4\lambda^{2}v^{2} + 1} - 1} \right)} \right)}\\
    &= {\log(n)} + \frac{d_{k}}{2}\left\lbrack {{\log\left( {\sqrt{4\lambda^{2}v^{2} + 1} - 1} \right)} - {\log\left( 2\lambda^{2}v^{2} \right)}} \right\rbrack - \frac{1}{2}{\log\sqrt{4\lambda^{2}v^{2} + 1}}
\end{align*}
When $\left. \lambda\rightarrow 0 \right.$, $\left. v\rightarrow 0 \right.$, we obtain $\left. {\log\left( \sqrt{4\lambda^{2}v^{2} + 1} \right)}\rightarrow 0 \right.$
\begin{align*}
    H_{i} \cong \widehat{H_{i}} &\cong {\mathit{\log}(n)} + \frac{d_{k}}{2}\left\lbrack {{\log\left( {\sqrt{4\lambda^{2}v^{2} + 1} - 1} \right)} - {\log\left( 2\lambda^{2}v^{2} \right)}} \right\rbrack\\
    &= {\mathit{\log}(n)} + \frac{d_{k}}{2}\left\lbrack \left. {\log\left( \frac{\left( {\sqrt{4\lambda^{2}v^{2} + 1} - 1} \right)}{2\lambda^{2}v^{2}} \right.} \right) \right\rbrack
\end{align*}
Let $t = \lambda^{2}v^{2}$,
\begin{align*}
    H_{i} \cong \widehat{H_{i}} \cong {\mathit{\log}(n)} + \frac{d_{k}}{2}\left\lbrack \left. {\log\left( \frac{\left( {\sqrt{4t + 1} - 1} \right)}{2t} \right.} \right) \right\rbrack
\end{align*}
Since $\left. \lambda\rightarrow 0 \right.$ and $\left. v\rightarrow 0 \right.$, we perform a Taylor expansion \cite{stewart2012calculus} of $\left. \sqrt{4t + 1} - 1 \right)$ at $t = 0$:
\begin{align*}
    \sqrt{4t + 1} - 1 &= 0 + \frac{2}{\sqrt{4t + 1}}\left. \hspace{0pt} \right|_{t = 0}\left( {t - 0} \right) + \frac{\left( {- 2} \right)}{\left( {4t + 1} \right)^{\frac{3}{2}}}\left. \hspace{0pt} \right|_{t = 0}\left( {t - 0} \right)^{2} + o\left( t^{3} \right)= 2t - 2t^{2} + o\left( t^{3} \right)
\end{align*}
Substituting Eq.\ref{eq11} into Eq. \ref{eq10} and we get:
\begin{align*}
    \widehat{H_{i}} \cong {\log(n)} + \frac{d_{k}}{2}\left\lbrack \left. {\log\left( \frac{\left( {2t - 2t^{2} + o\left( t^{3} \right)} \right)}{2t} \right.} \right) \right\rbrack \cong {\log(n)} + \frac{d_{k}}{2}\left\lbrack \left. {\log\left( 1 - t \right.} \right) \right\rbrack = \epsilon
\end{align*}
where $\epsilon$ is a hyperparameter that we wish the information entropy of the attention score will be constant regardless the input length $n$.
\begin{align*}
    \lambda = \frac{\sqrt{1 - e^{\frac{2\epsilon}{d_{k}}}n^{- \frac{2}{d_{k}}}}}{v} = \mathcal{K}\sqrt{1 - e^{\frac{2\epsilon}{d_{k}}}n^{- \frac{2}{d_{k}}}}
\end{align*}
When the model was pre-trained by maximum length $n_{tr}$ with scaler $\frac{1}{\sqrt{d_{k}}}$, which is $n=n_{tr}$, we suppose the model was perfectly trained, then we have:
\begin{align*}
    \frac{1}{\sqrt{d_{k}}} = \lambda = \mathcal{K}\sqrt{1 - e^{\frac{2\epsilon}{d_{k}}}{n_{tr}}^{- \frac{2}{d_{k}}}},\mathcal{K} = \frac{1}{\sqrt{\left( {1 - e^{\frac{2\epsilon}{d_{k}}}{n_{tr}}^{- \frac{2}{d_{k}}}} \right)d_{k}}}
\end{align*}
Then we replace $\mathcal{K}$ and get:
\begin{align*}
    \lambda = \frac{\sqrt{1 - e^{\frac{2\epsilon}{d_{k}}}n^{- \frac{2}{d_{k}}}}}{\sqrt{\left( {1 - e^{\frac{2\epsilon}{d_{k}}}{n_{tr}}^{- \frac{2}{d_{k}}}} \right)d_{k}}} = \frac{\sqrt{1 - e^{\frac{2\epsilon}{d_{k}}}n^{- \frac{2}{d_{k}}}}}{\sqrt{\left( {1 - e^{\frac{2\epsilon}{d_{k}}}{n_{tr}}^{- \frac{2}{d_{k}}}} \right)}}*\frac{1}{\sqrt{d_{k}}} = \frac{InfoScale}{\sqrt{d_{k}}}
\end{align*}
Finally, we call the scaled temperature of the scaler $\sqrt{\dfrac{1 - e^{\frac{2\epsilon}{d_{k}}}n^{- \frac{2}{d_{k}}}}{1 - e^{\frac{2\epsilon}{d_{k}}}{n_{tr}}^{- \frac{2}{d_{k}}}}}$ as InfoScale. 
\subsection{Detailed Derivations of CosScale}
\label{The details of the deduction of CosScale}
\textbf{\cref{theo1}}
The peak value $\eta_{1}^{*}$ of the QK distribution before applying RoPE progressively shifts towards 1 as CosScale $\alpha$ increases.
     
Recall that GAU-α applies RoPE to the query-key product (QK) before the softmax normalization. Therefore, we have $a_{ij} = e^{\alpha{\cos\theta_{ij}}}$ and its expectation as:
\begin{align*}
    \mathbb{E}\left\lbrack a_{ij} \right\rbrack = \frac{1}{n^{2}}{\sum_{ij}^{n^{2}}e^{\alpha{\cos\theta_{ij}}}} \cong \mathbb{E}\left\lbrack e^{\alpha{\cos\theta}} \right\rbrack
\end{align*}
Consider the probability density function of a \textit{n}-dimension hypersphere:
\begin{align*}
    p_{n}(\theta) = \frac{\Gamma\left( \frac{n}{2} \right)}{\Gamma\left( \frac{n - 1}{2} \right)\sqrt{\pi}}{\sin^{n - 2}(\theta)}
\end{align*}
For the scaled cosine attention, let $\eta = {\cos\theta}$ and then we have:
\begin{align*}
    p_{d_k}(\eta) &= \frac{\Gamma\left( \frac{d_{k}}{2} \right)}{\Gamma\left( \frac{d_{k} - 1}{2} \right)\sqrt{\pi}}\left( {1 - \eta^{2}} \right)^{\frac{d_{k} - 3}{2}}\\
    \mathbb{E}\left\lbrack a_{ij} \right\rbrack \cong \mathbb{E}\left\lbrack e^{\alpha{\cos\theta}} \right\rbrack &= \frac{{\int_{- 1}^{1}{e^{\alpha\eta}\left( {1 - \eta^{2}} \right)^{\frac{d_{k} - 3}{2}}}}\mathrm{d}\eta}{{\int_{- 1}^{1}\left( {1 - \eta^{2}} \right)^{\frac{d_{k} - 3}{2}}}\mathrm{d}\eta} = \frac{\boldsymbol{{I}_{1}}}{\boldsymbol{{I}_{0}}}
\end{align*}
It turns out that the exponential functions $e^{\alpha\eta}$ makes the integrals $\boldsymbol{{I}_{1}}$ have the maximum value if $\left. \eta\rightarrow 1 \right.$ when $\alpha$ increases.
\begin{gather*}
    p_{1}^{'}(\eta) = \alpha e^{\alpha\eta}\left( {1 - \eta^{2}} \right)^{\frac{d_{k} - 3}{2}} - \left( {d_{k} - 3} \right)\eta e^{\alpha\eta}\left( {1 - \eta^{2}} \right)^{\frac{d_{k} - 5}{2}} = 0\\
    \alpha\left( {1 - \eta^{2}} \right) - \left( {d_{k} - 3} \right)\eta = 0\\
    \eta_{1}^{*} = \frac{- \left( {d_{k} - 3} \right) + \sqrt{\left( {d_{k} - 3} \right)^{2} + 4\alpha^{2}}}{2\alpha}
\end{gather*}
\subsection{Validation of CosScale Theorems}
\label{Testify the validation of the CosScale's Theorems}
\subsubsection{Validation of Theorem 1}
\begin{figure}[]
\vskip 0.1in
\begin{center}
\centerline{\includegraphics[width=0.6\textwidth]{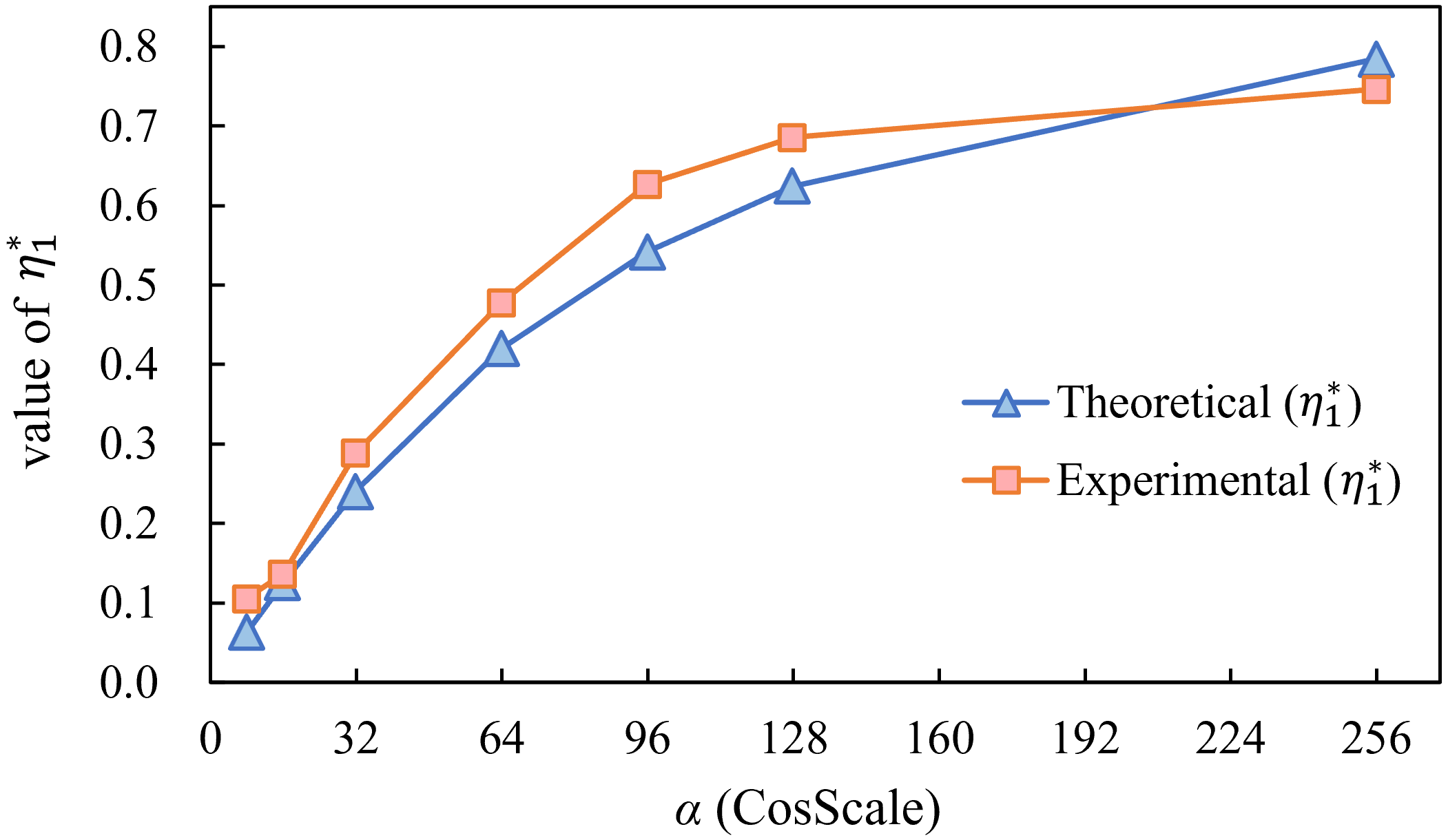}}
\caption{The comparison of theoretical $\eta_{1}^{*}$ and experimental $\eta_{1}^{*}$ at different $\alpha$ values according to Eq. \ref{eq18}. The x-axis represents $\alpha$ (CosScale), and the y-axis represents the peak value of $\eta_{1}^{*}$.}
\label{f6}
\end{center}
\vskip -0.1in
\end{figure}

To validate \cref{theo1}, we compare the theoretical peak values of $\eta_{1}^{*}$ predicted by Eq. \ref{eq18} with the corresponding experimental values obtained by varying CosScale from 8 to 256. \cref{f6} shows the close agreement between the theoretical and experimental results, confirming the validity of \cref{theo1} for all CosScale values.
\subsubsection{Validation of Theorem 2}
\begin{figure}[]
\vskip 0.1in
\begin{center}
\centerline{\includegraphics[width=\textwidth]{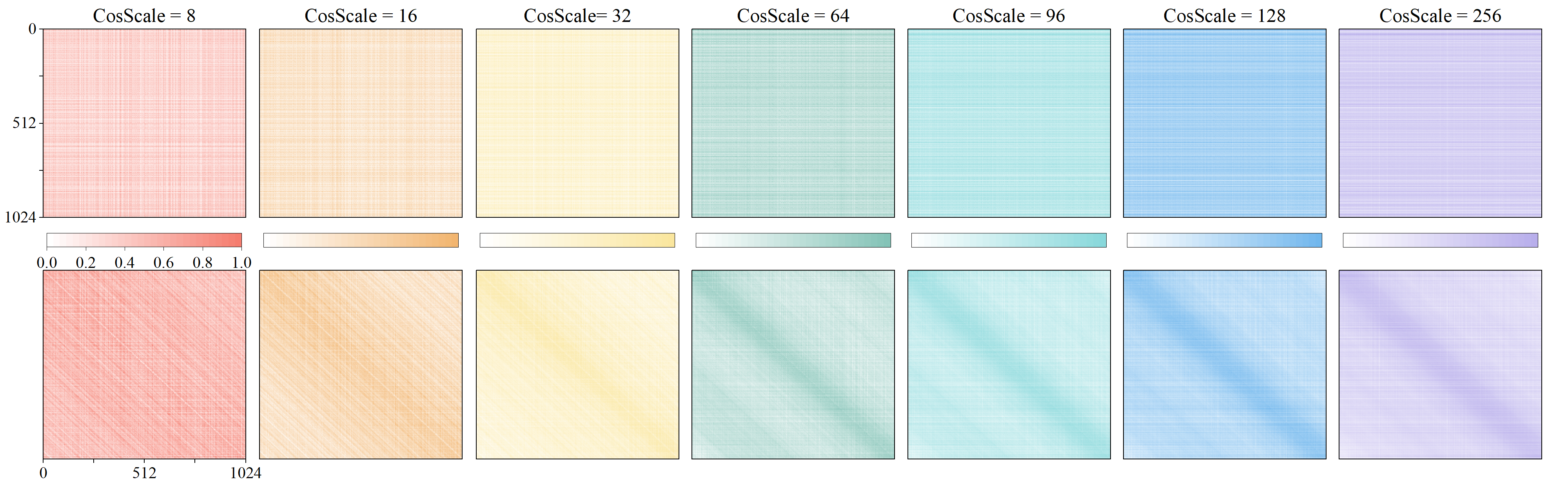}}
\caption{Heatmaps of QK multiplications normalized by global maximum and minimum values before and after RoPE processing(range from 0 to 1) with extending sequence length to 1024. The first row represents the heatmaps of QK multiplication before RoPE at increasing CosScale values (from left to right: 8, 16, 32, 64, 96, 128, 256). And the second row shows the corresponding heatmaps of QK multiplication after RoPE at increasing CosScale values (from left to right: 8, 16, 32, 64, 96, 128, 256). Each pair of two heatmaps in a column shares a consistent color bar between them.}
\label{f7}
\end{center}
\vskip -0.1in
\end{figure}

\cref{f7} provides visual evidence supporting Theorem 2. Taking sequence length of 1024 as an example, as the CosScale value increases from 8 to 256, we observe an increasingly distinct horizontal stripe pattern in the heatmaps of the QK multiplication (\cref{f7} (a)). This pattern translates to a progressively clearer RoPE-type attention score pattern after applying RoPE (\cref{f7}(b)). These observations align with the predictions of Theorem 2, demonstrating the impact of CosScale on the structure of the attention matrix.
\subsection{Performance of Published Methods on GAU-α}
\label{All published methods fail on GAU-α}
\cref{table 9.7} presents the performance of eight published length extrapolation methods on the GAU-α architecture. We observe that these methods generally exhibit different behavior on GAU-α compared to their reported performance on LLaMA.

For RoPE-based methods: YaRN extrapolates high-frequency signals and interpolates low-frequency signals. YaRN achieves the best performance among the three methods, but its perplexity remains higher than that reported for LLaMA (reported PPL = 3.5 in YaRN[4]). ReRoPE interpolates low-frequency signals while truncates high-frequency signals, resulting in inferior performance compared to YaRN. PI performs signal interpolation across all frequencies, hindering the distinction of low-frequency signals and leading to the worst performance.

The bias method ALiBi performs poorly, possibly due to the Post-LN structure of GAU-α \cite{wang2024deepnet}, which increases model depth and may reduce the effectiveness of ALiBi. This may also explain the suboptimal performance of the skip-wise training method, PoSE.

For window-based methods: the Windowed Attention achieves the best performance among the three. StreamingLLM places extra emphasis on initial tokens, which may not always be semantically significant \cite{xiao2023efficient}. This, combined with the truncation of high-frequency signals by ReRoPE, may contribute to its lower performance. LM-Infinite combines ReRoPE and Λ-shaped attention, with a focus on initial tokens. However, its performance is hindered by the lack of "attention sink" behavior in our experiments, suggesting that the attention mechanism in GAU-α differs from the models used in \cite{han2024lm}.

These findings highlight the importance of considering model architecture and attention mechanisms when evaluating length extrapolation methods. The varying performance of these methods on GAU-α underscores the need for further research to develop more robust and generalizable techniques.
\begin{table*}[]
\caption{Perplexity (PPL) and accuracy (ACC) for baseline methods evaluated on GAU-α with sequence lengths ranging from 128 to 4k on the length extrapolation task. Lower PPL indicates better performance, while higher ACC is preferable.}
\label{table 9.7}
\centering
\vskip 0.1in
\begin{tabular}{l@{\hspace{0.1cm}}|c@{\hspace{0.15cm}}c@{\hspace{0.1cm}}l@{\hspace{0.25cm}}c@{\hspace{0.15cm}}c@{\hspace{0.1cm}}l@{\hspace{0.25cm}}c@{\hspace{0.15cm}}c@{\hspace{0.1cm}}l@{\hspace{0.25cm}}c@{\hspace{0.15cm}}c@{\hspace{0.1cm}}l@{\hspace{0.25cm}}c@{\hspace{0.15cm}}c@{\hspace{0.1cm}}l@{\hspace{0.25cm}}c@{\hspace{0.15cm}}c@{\hspace{0.1cm}}}
\toprule
\multirow{3}{*}{Methods} & \multicolumn{17}{c}{\textbf{Evaluation Length}}                                                                                                                                                                                    \\ \cline{2-18} 
                         & \multicolumn{2}{c}{128}        &  & \multicolumn{2}{c}{256}        &  & \multicolumn{2}{c}{512}            &  & \multicolumn{2}{c}{1024}           &  & \multicolumn{2}{c}{2048}           &  & \multicolumn{2}{c}{4096}           \\ \cline{2-3} \cline{5-6} \cline{8-9} \cline{11-12} \cline{14-15} \cline{17-18} 
                         & PPL           & ACC.           &  & PPL           & ACC.           &  & PPL               & ACC.           &  & PPL               & ACC.           &  & PPL               & ACC.           &  & PPL               & ACC.           \\ \hline
PI                       & 24.51         & 0.46           &  & 10.75         & 0.55           &  & \textgreater{}500 & \textless{}0.1 &  & \textgreater{}500 & \textless{}0.1 &  & \textgreater{}500 & \textless{}0.1 &  & \textgreater{}500 & \textless{}0.1 \\
ALiBi                    & 151.00        & \textless{}0.1 &  & 271.97        & \textless{}0.1 &  & 346.88            & \textless{}0.1 &  & 401.87            & \textless{}0.1 &  & 488.47            & \textless{}0.1 &  & \textgreater{}500 & \textless{}0.1 \\
PoSE                     & \textbf{1.97} & \textbf{0.85}  &  & \textbf{2.23} & \textbf{0.81}  &  & \textbf{2.61}     & 0.77           &  & 14.74             & 0.46           &  & 240.25            & \textless{}0.1 &  & \textgreater{}500 & \textless{}0.1 \\
StreamingLLM             & 2.18          & 0.84           &  & 2.55          & 0.80           &  & 2.85              & 0.78           &  & 4.27              & 0.71           &  & 4.88              & 0.69           &  & 5.42              & 0.67           \\
YaRN(s=16)               & 2.41          & 0.81           &  & 2.99          & 0.76           &  & 3.37              & 0.73           &  & 7.07              & 0.60           &  & 15.34             & 0.48           &  & 26.59             & 0.41           \\
YaRN(s=32)               & 2.65          & 0.79           &  & 3.35          & 0.74           &  & 3.88              & 0.71           &  & 7.94              & 0.58           &  & 16.68             & 0.47           &  & 27.59             & 0.40           \\
W.A.                     & 2.19          & 0.84           &  & 2.55          & 0.80           &  & 2.82              & \textbf{0.78}  &  & \textbf{3.69}     & \textbf{0.74}  &  & \textbf{3.87}     & \textbf{0.73}  &  & \textbf{4.24}     & \textbf{0.71}  \\
LM-Infinite              & 12.25         & 0.55           &  & 18.65         & 0.46           &  & 21.66             & 0.43           &  & 30.72             & 0.39           &  & 32.04             & 0.38           &  & 32.56             & 0.37           \\
ReRoPE                   & 15.61         & 0.49           &  & 46.44         & 0.33           &  & 151.22            & 0.18           &  & 295.42            & 0.12           &  & 475.82            & \textless{}0.1 &  & \textgreater{}500 & \textless{}0.1\\
\bottomrule
\end{tabular}
\end{table*}
\subsection{Additional Tables and Charts}
\label{Additional Tables and Charts}
\begin{table*}[t]
\caption{Summary of the four related methods. The name of each method, the function of the scaled dot-product, and the corresponding scaled temperature are listed here.}
\label{table1}
\vskip 0.1in
\begin{center}
\begin{small}
\begin{tabular}{c|cc}
\toprule
Methods & Scaled dot-product & Scale Temperature \\
\midrule
Vanilla Transformer & \(\displaystyle softmax( \frac{1}{\sqrt{d_k}} QK^T )\) & \(1\) \\
\\[0.1pt]
Log-length Scaling & \(\displaystyle softmax ( \frac{\mathit{\log}~n_{te}}{\sqrt{d_{k}}}{QK}^{T} )\) & \(\mathit{\log}~n_{te}\) \\
\\[0.1pt]
Softmax Plus & \(\displaystyle softmax( \frac{\mathit{\log}_{512}~n_{te}}{\sqrt{d_{k}}}{QK}^{T} )\) & \(\mathit{\log}_{512}~n_{te}\)\\
\\[0.1pt]
Pre-softmax Scaling & \(\displaystyle 
softmax\left( {\frac{1}{t\sqrt{d_{k}}}{QK}^{T}} \right),~\sqrt{\frac{1}{t}} = 0.1~{\mathit{\ln}\left( \frac{n_{te}}{n_{tr}} \right)} + 1~\) & \(\displaystyle \left( {0.1~{\mathit{\ln}\left( \frac{n_{te}}{n_{tr}} \right)} + 1} \right)^{2}\) \\
\\[0.1pt]
\textbf{InfoScale (Ours)} & \(\bm{\displaystyle softmax\left( {\frac{InfoScale}{\sqrt{d_{k}}}{QK}^{T}} \right),InfoScale = \sqrt{\frac{{1 - n_{te}^{- \frac{2}{d_k}}}}{ {1 - {n_{tr}}^{- \frac{2}{d_k}}} }}}\) & \(\bm{\displaystyle \sqrt{\frac{{1 - n_{te}^{- \frac{2}{d_k}}}}{ {1 - {n_{tr}}^{- \frac{2}{d_k}}}}} }\)\\
\bottomrule
\end{tabular}
\end{small}
\end{center}
\vskip -0.1in
\end{table*}

\begin{table*}[t]
\caption{Perplexity (PPL) and accuracy (ACC) for baseline methods and their counterparts with InfoScale incorporated, evaluated on the GAU-α model with sequence lengths ranging from 128 to 4k on the length extrapolation task. For simplicity, W.A. denotes the baseline algorithm the Windowed Attention, and I.S. represents our method InfoScale. }
\label{table 4.2.1}
\vskip 0.1in
\begin{center}
\begin{tabular}{l@{\hspace{0.1cm}}|c@{\hspace{0.1cm}}c@{\hspace{0.1cm}}l@{\hspace{0.25cm}}c@{\hspace{0.1cm}}c@{\hspace{0.1cm}}l@{\hspace{0.25cm}}c@{\hspace{0.1cm}}c@{\hspace{0.1cm}}l@{\hspace{0.25cm}}c@{\hspace{0.1cm}}c@{\hspace{0.1cm}}l@{\hspace{0.25cm}}c@{\hspace{0.1cm}}c@{\hspace{0.1cm}}l@{\hspace{0.25cm}}c@{\hspace{0.1cm}}c@{\hspace{0.1cm}}}
\toprule
\multirow{3}{*}{Methods} & \multicolumn{17}{c}{Evaluation   Length}                                                                                                                                                                                                      \\ \cline{2-18} 
                         & \multicolumn{2}{c}{128}         &  & \multicolumn{2}{c}{256}         &  & \multicolumn{2}{c}{512}            &  & \multicolumn{2}{c}{1024}           &  & \multicolumn{2}{c}{2048}           &  & \multicolumn{2}{c}{4096}                    \\ \cline{2-3} \cline{5-6} \cline{8-9} \cline{11-12} \cline{14-15} \cline{17-18} 
                         & PPL           & ACC.           &  & PPL           & ACC.           &  & PPL              & ACC.           &  & PPL              & ACC.           &  & PPL              & ACC.           &  & PPL              & ACC.                    \\ \hline
PI                       & 24.51          & 0.46           &  & 10.75          & 0.55           &  & \textgreater{}500 & \textless{}0.1 &  & \textgreater{}500 & \textless{}0.1 &  & \textgreater{}500 & \textless{}0.1 &  & \textgreater{}500 & \textless{}0.1          \\
PI w/ I.S.               & 21.67          & 0.48           &  & 8.14           & 0.61           &  & \textgreater{}500 & \textless{}0.1 &  & \textgreater{}500 & \textless{}0.1 &  & \textgreater{}500 & \textless{}0.1 &  & \textgreater{}500 & \textless{}0.1          \\ \hline
ALiBi                    & 151.00         & \textless{}0.1 &  & 271.97         & \textless{}0.1 &  & 346.88            & \textless{}0.1 &  & 401.87            & \textless{}0.1 &  & 488.47            & \textless{}0.1 &  & \textgreater{}500 & \textless{}0.1          \\
ALiBi w/ I.S.            & 150.74         & \textless{}0.1 &  & 270.76         & \textless{}0.1 &  & 353.25            & \textless{}0.1 &  & 404.16            & \textless{}0.1 &  & 478.59            & \textless{}0.1 &  & \textgreater{}500 & \textless{}0.1          \\ \hline
PoSE                     & 1.97           & 0.85           &  & 2.23           & 0.81           &  & 2.61              & 0.77           &  & 14.74             & 0.46           &  & 240.25            & 0.10           &  & \textgreater{}500 & \textless{}0.1          \\
PoSE w/ I.S.             & \textbf{1.95}  & \textbf{0.85}  &  & \textbf{2.10}  & \textbf{0.83}  &  & \textbf{2.15}     & \textbf{0.82}  &  & \textbf{6.32}     & \textbf{0.60}  &  & \textbf{73.80}    & \textbf{0.23}  &  & \textbf{354.50}   & \textbf{\textless{}0.1} \\ \hline
StreamingLLM             & 2.18           & \textbf{0.84}  &  & 2.55           & 0.80           &  & 2.85              & 0.78           &  & \textbf{4.27}     & \textbf{0.71}  &  & 4.88              & 0.69           &  & 5.42              & 0.67                    \\
StreamingLLM w/ I.S.     & \textbf{2.18}  & 0.83           &  & \textbf{2.53}  & \textbf{0.80}  &  & \textbf{2.85}     & \textbf{0.79}  &  & 4.76              & 0.70           &  & \textbf{4.84}     & \textbf{0.70}  &  & \textbf{5.20}     & \textbf{0.68}           \\ \hline
YaRN(s=16)               & 2.41           & 0.81           &  & 2.99           & 0.76           &  & 3.37              & 0.73           &  & 7.07              & 0.60           &  & 15.34             & 0.48           &  & 26.59             & 0.41                    \\
YaRN(s=16) w/ I.S.       & \textbf{2.26}  & \textbf{0.82}  &  & \textbf{2.58}  & \textbf{0.80}  &  & \textbf{2.61}     & \textbf{0.78}  &  & \textbf{4.37}     & \textbf{0.70}  &  & \textbf{7.49}     & \textbf{0.61}  &  & \textbf{12.38}    & \textbf{0.53}           \\ \hline
YaRN(s=32)               & 2.65           & 0.79           &  & 3.35           & 0.74           &  & 3.88              & 0.71           &  & 7.94              & 0.58           &  & 16.68             & 0.47           &  & 27.59             & 0.40                    \\
YaRN(s=32) w/ I.S.       & \textbf{2.42}  & \textbf{0.81}  &  & \textbf{2.74}  & \textbf{0.78}  &  & \textbf{2.75}     & \textbf{0.77}  &  & \textbf{4.40}     & \textbf{0.69}  &  & \textbf{7.04}     & \textbf{0.62}  &  & \textbf{10.56}    & \textbf{0.55}           \\ \hline
W.A.                     & 2.19           & \textbf{0.84}  &  & 2.55           & 0.80           &  & \textbf{2.82}     & 0.78           &  & \textbf{3.69}     & 0.74           &  & \textbf{3.87}     & 0.73           &  & \textbf{4.24}     & 0.71                    \\
W.A. w/ I.S.             & \textbf{2.19}  & 0.83           &  & \textbf{2.53}  & \textbf{0.80}  &  & 2.83              & \textbf{0.79}  &  & 3.73              & \textbf{0.74}  &  & 3.92              & \textbf{0.74}  &  & 4.32              & \textbf{0.72}           \\ \hline
LM-Infinite              & \textbf{12.25} & \textbf{0.55}  &  & \textbf{18.65} & \textbf{0.46}  &  & \textbf{21.66}    & \textbf{0.43}  &  & \textbf{30.72}    & \textbf{0.39}  &  & 32.04             & 0.38           &  & \textbf{32.56}    & 0.37                    \\
LM-Infinite w/ I.S.      & 12.66          & 0.54           &  & 26.64          & 0.42           &  & 28.47             & 0.40           &  & 30.90             & 0.38           &  & \textbf{29.92}    & \textbf{0.38}  &  & 32.81             & \textbf{0.37}           \\ \hline
ReRoPE                   & 15.61          & 0.49           &  & 46.44          & 0.33           &  & 151.22            & 0.18           &  & 295.42            & 0.12           &  & 475.82            & \textless{}0.1 &  & \textgreater{}500 & \textless{}0.1          \\
ReRoPE w/ I.S.           & \textbf{14.77} & \textbf{0.50}  &  & \textbf{23.61} & \textbf{0.43}  &  & \textbf{52.86}    & \textbf{0.32}  &  & \textbf{115.83}   & \textbf{0.23}  &  & \textbf{210.78}   & \textbf{0.17}  &  & \textbf{311.24}   & \textbf{0.12} \\
\bottomrule
\end{tabular}
\end{center}
\end{table*}

\begin{table*}[t]
\caption{Perplexity (PPL) and accuracy (ACC) for baseline methods and their counterparts with CosScale incorporated, evaluated with sequence lengths ranging from 128 to 4k on the length extrapolation task. Lower PPL and higher ACC indicate better performance. CosScale is set to 128 for all methods except the Windowed Attention, where it is set to 16. W.A. denotes the Windowed Attention, and C.S. represents CosScale.}
\label{table 4.2.2}
\vskip 0.1in
\begin{center}
\begin{tabular}{l@{\hspace{0.1cm}}|c@{\hspace{0.1cm}}c@{\hspace{0.1cm}}l@{\hspace{0.25cm}}c@{\hspace{0.1cm}}c@{\hspace{0.1cm}}l@{\hspace{0.25cm}}c@{\hspace{0.1cm}}c@{\hspace{0.1cm}}l@{\hspace{0.25cm}}c@{\hspace{0.1cm}}c@{\hspace{0.1cm}}l@{\hspace{0.25cm}}c@{\hspace{0.1cm}}c@{\hspace{0.1cm}}l@{\hspace{0.25cm}}c@{\hspace{0.1cm}}c@{\hspace{0.1cm}}}
\toprule
\multirow{3}{*}{Methods} & \multicolumn{17}{c}{Evaluation   Length}                                                                                                                                                                                                                                        \\ \cline{2-18} 
                         & \multicolumn{2}{c}{128}        &           & \multicolumn{2}{c}{256}        &           & \multicolumn{2}{c}{512}            &           & \multicolumn{2}{c}{1024}           &           & \multicolumn{2}{c}{2048}           &           & \multicolumn{2}{c}{4096}           \\ \cline{2-3} \cline{5-6} \cline{8-9} \cline{11-12} \cline{14-15} \cline{17-18} 
                         & PPL           & ACC.           &           & PPL           & ACC.           &           & PPL               & ACC.           &           & PPL               & ACC.           &           & PPL               & ACC.           &           & PPL               & ACC.           \\ \hline
PI                       & 24.51         & 0.46           &           & 10.75         & 0.55           &           & \textgreater{}500 & \textless{}0.1 &           & \textgreater{}500 & \textless{}0.1 &           & \textgreater{}500 & \textless{}0.1 &           & \textgreater{}500 & \textless{}0.1 \\
PI w/ C.S.               & 10.38         & 0.52           &           & 10.46         & 0.53           &           & \textgreater{}500 & \textless{}0.1 &           & \textgreater{}500 & \textless{}0.1 &           & \textgreater{}500 & \textless{}0.1 &           & \textgreater{}500 & \textless{}0.1 \\ \hline
ALiBi                    & 151.00        & \textless{}0.1 &           & 271.97        & \textless{}0.1 &           & 346.88            & \textless{}0.1 &           & 401.87            & \textless{}0.1 &           & 488.47            & \textless{}0.1 &           & \textgreater{}500 & \textless{}0.1 \\
ALiBi w/ C.S.            & 151.57        & \textless{}0.1 &           & 263.90        & \textless{}0.1 &           & 335.35            & \textless{}0.1 &           & 392.22            & \textless{}0.1 &           & 468.78            & \textless{}0.1 &           & 486.49            & \textless{}0.1 \\ \hline
PoSE                     & \textbf{1.97} & \textbf{0.85}  & \textbf{} & \textbf{2.23} & 0.81           &           & 2.61              & 0.77           &           & 14.74             & 0.46           &           & 240.25            & 0.10           &           & \textgreater{}500 & \textless{}0.1 \\
PoSE w/ C.S.             & 2.07          & 0.84           &           & 2.29          & \textbf{0.81}  & \textbf{} & \textbf{2.34}     & \textbf{0.81}  & \textbf{} & \textbf{3.35}     & \textbf{0.73}  & \textbf{} & \textbf{12.37}    & \textbf{0.51}  & \textbf{} & \textbf{22.03}    & \textbf{0.41}  \\ \hline
StreamingLLM             & \textbf{2.18} & \textbf{0.84}  & \textbf{} & \textbf{2.55} & \textbf{0.80}  & \textbf{} & \textbf{2.85}     & \textbf{0.78}  & \textbf{} & 4.27              & 0.71           &           & 4.88              & 0.69           &           & 5.42              & 0.67           \\
StreamingLLM w/ C.S.     & 2.31          & 0.82           &           & 2.72          & 0.78           &           & 2.97              & 0.77           &           & \textbf{4.13}     & \textbf{0.72}  & \textbf{} & \textbf{4.59}     & \textbf{0.70}  & \textbf{} & \textbf{4.88}     & \textbf{0.69}  \\ \hline
YaRN(s=16)               & \textbf{2.41} & \textbf{0.81}  & \textbf{} & \textbf{2.99} & \textbf{0.76}  & \textbf{} & 3.37              & 0.73           &           & 7.07              & 0.60           &           & 15.34             & 0.48           &           & 26.59             & 0.41           \\
YaRN(s=16) w/ C.S.       & 2.57          & 0.78           &           & 3.09          & 0.75           &           & \textbf{3.09}     & \textbf{0.75}  & \textbf{} & \textbf{4.79}     & \textbf{0.67}  & \textbf{} & \textbf{6.28}     & \textbf{0.63}  & \textbf{} & \textbf{7.73}     & \textbf{0.59}  \\ \hline
YaRN(s=32)               & \textbf{2.65} & \textbf{0.79}  & \textbf{} & 3.35          & \textbf{0.74}  & \textbf{} & 3.88              & 0.71           &           & 7.94              & 0.58           &           & 16.68             & 0.47           &           & 27.59             & 0.40           \\
YaRN(s=32) w/ C.S.       & 2.70          & 0.78           &           & \textbf{3.29} & 0.73           &           & \textbf{3.40}     & \textbf{0.73}  & \textbf{} & \textbf{5.29}     & \textbf{0.65}  & \textbf{} & \textbf{6.87}     & \textbf{0.61}  & \textbf{} & \textbf{8.50}     & \textbf{0.58}  \\ \hline
W.A.                     & \textbf{2.19} & 0.84           &           & \textbf{2.55} & \textbf{0.80}  & \textbf{} & \textbf{2.82}     & \textbf{0.78}  & \textbf{} & \textbf{3.69}     & \textbf{0.74}  & \textbf{} & \textbf{3.87}     & \textbf{0.73}  & \textbf{} & \textbf{4.24}     & \textbf{0.71}  \\
W.A.(C.S.=16) w/ C.S.             & 2.21          & \textbf{0.84}  & \textbf{} & 2.67          & 0.79           &           & 2.93              & 0.78           &           & 3.81              & 0.73           &           & 4.02              & 0.73           &           & 4.44              & 0.71           \\ \hline
LM-Infinite              & 12.25         & 0.55           &           & 18.65         & 0.46           &           & 21.66             & 0.43           &           & 30.72             & 0.39           &           & 32.04             & 0.38           &           & 32.56             & 0.37           \\
LM-Infinite w/ C.S.      & \textbf{4.09} & \textbf{0.69}  & \textbf{} & \textbf{6.63} & \textbf{0.61}  & \textbf{} & \textbf{7.99}     & \textbf{0.58}  & \textbf{} & \textbf{10.88}    & \textbf{0.54}  & \textbf{} & \textbf{11.83}    & \textbf{0.53}  & \textbf{} & \textbf{12.91}    & \textbf{0.50}  \\ \hline
ReRoPE                   & 15.61         & 0.49           &           & 46.44         & 0.33           &           & 151.22            & 0.18           &           & 295.42            & 0.12           &           & 475.82            & \textless{}0.1 &           & \textgreater{}500 & \textless{}0.1 \\
ReRoPE w/ C.S.           & \textbf{3.69} & \textbf{0.72}  & \textbf{} & \textbf{3.62} & \textbf{0.72}  & \textbf{} & \textbf{3.88}     & \textbf{0.71}  & \textbf{} & \textbf{5.10}     & \textbf{0.67}  & \textbf{} & \textbf{5.55}     & \textbf{0.66}  & \textbf{} & \textbf{6.36}     & \textbf{0.63} \\
\bottomrule
\end{tabular}   
\end{center}
\end{table*}

\begin{table*}[t]
\caption{Validation performance across different CosScale values with cosine attention. Lower PPL indicates better performance, while higher ACC is preferable.}
\label{table 4.3.2}
\centering
\vskip 0.1in
\begin{tabular}{c@{\hspace{0.1cm}}|c@{\hspace{0.15cm}}c@{\hspace{0.1cm}}l@{\hspace{0.25cm}}c@{\hspace{0.15cm}}c@{\hspace{0.1cm}}l@{\hspace{0.25cm}}c@{\hspace{0.15cm}}c@{\hspace{0.1cm}}l@{\hspace{0.25cm}}c@{\hspace{0.15cm}}c@{\hspace{0.1cm}}l@{\hspace{0.25cm}}c@{\hspace{0.15cm}}c@{\hspace{0.1cm}}l@{\hspace{0.25cm}}c@{\hspace{0.15cm}}c@{\hspace{0.1cm}}}
\toprule
\multirow{3}{*}{CosScale} & \multicolumn{17}{c}{Evaluation Length}                                                                                                                                                                                      \\ \cline{2-18} 
                          & \multicolumn{2}{c}{128}       &  & \multicolumn{2}{c}{256}       &  & \multicolumn{2}{c}{512}       &  & \multicolumn{2}{c}{1024}           &  & \multicolumn{2}{c}{2048}           &  & \multicolumn{2}{c}{4096}           \\ \cline{2-3} \cline{5-6} \cline{8-9} \cline{11-12} \cline{14-15} \cline{17-18} 
                          & PPL           & ACC.          &  & PPL           & ACC.          &  & PPL           & ACC.          &  & PPL               & ACC.           &  & PPL               & ACC.           &  & PPL               & ACC.           \\ \hline
8                         & 2.86          & 0.77          &  & 16.14         & 0.46          &  & 237.15        & 0.11          &  & \textgreater{}500 & \textless{}0.1 &  & \textgreater{}500 & \textless{}0.1 &  & \textgreater{}500 & \textless{}0.1 \\
16                        & 2.30          & 0.83          &  & 4.38          & 0.70          &  & 30.16         & 0.37          &  & 320.84            & 0.12           &  & \textgreater{}500 & \textless{}0.1 &  & \textgreater{}500 & \textless{}0.1 \\
32                        & \textbf{2.17} & \textbf{0.84} &  & 2.73          & 0.78          &  & 4.63          & 0.68          &  & 76.40             & 0.26           &  & 499.79            & \textless{}0.1 &  & \textgreater{}500 & \textless{}0.1 \\
64                        & 2.21          & 0.82          &  & \textbf{2.60} & \textbf{0.78} &  & 2.84          & 0.77          &  & 5.73              & 0.64           &  & 42.32             & 0.34           &  & 267.49            & 0.13           \\
96                        & 2.23          & 0.82          &  & 2.63          & 0.78          &  & \textbf{2.83} & \textbf{0.77} &  & 5.12              & 0.67           &  & 25.20             & 0.42           &  & 65.72             & 0.29           \\
128                       & 2.29          & 0.82          &  & 2.70          & 0.78          &  & 2.92          & 0.77          &  & 5.03              & 0.68           &  & 24.03             & 0.43           &  & \textbf{49.45}    & \textbf{0.32}  \\
256                       & 2.65          & 0.78          &  & 3.23          & 0.74          &  & 3.48          & 0.74          &  & \textbf{4.72}     & \textbf{0.69}  &  & 18.34             & 0.49           &  & 60.67             & 0.31           \\
600                       & 3.82          & 0.72          &  & 4.61          & 0.69          &  & 4.80          & 0.68          &  & 6.14              & 0.64           &  & \textbf{13.63}    & \textbf{0.53}  &  & 79.33             & 0.28          \\
\bottomrule
\end{tabular}
\end{table*}

\begin{table*}[t]
\caption{Performance of InfoScale and CosScale on standard benchmarks. Lower PPL indicates better performance, while higher ACC is preferable. The best value for each extrapolation length and metric is highlighted in \textbf{bold}, while the second-best is indicated in \underline{underline}.}
\label{table 4.5}
\centering
\vskip 0.1in
\begin{tabular}{l@{\hspace{0.1cm}}|c@{\hspace{0.15cm}}c@{\hspace{0.1cm}}l@{\hspace{0.25cm}}c@{\hspace{0.15cm}}c@{\hspace{0.1cm}}l@{\hspace{0.25cm}}c@{\hspace{0.15cm}}c@{\hspace{0.1cm}}l@{\hspace{0.25cm}}c@{\hspace{0.15cm}}c@{\hspace{0.1cm}}l@{\hspace{0.25cm}}c@{\hspace{0.15cm}}c@{\hspace{0.1cm}}l@{\hspace{0.25cm}}c@{\hspace{0.15cm}}c@{\hspace{0.1cm}}}
\toprule
\multirow{3}{*}{Methods} & \multicolumn{17}{c}{Evaluation Length}                                                                                                                                                                             \\ \cline{2-18} 
                         & \multicolumn{2}{c}{128}       &  & \multicolumn{2}{c}{256}       &  & \multicolumn{2}{c}{512}       &  & \multicolumn{2}{c}{1024}      &  & \multicolumn{2}{c}{2048}       &  & \multicolumn{2}{c}{4096}           \\ \cline{2-3} \cline{5-6} \cline{8-9} \cline{11-12} \cline{14-15} \cline{17-18} 
                         & PPL           & ACC.          &  & PPL           & ACC.          &  & PPL           & ACC.          &  & PPL           & ACC.          &  & PPL            & ACC.          &  & PPL               & ACC.           \\ \hline
GAU-α                    & \underline{2.17}          & \textbf{0.84} &  & \underline{2.55}          & \underline{0.80}          &  & 3.05          & 0.76          &  & 17.12         & 0.46          &  & 327.44         & 0.10          &  & \textgreater{}500 & \textless{}0.1 \\
GAU-α   w/ C.S.          & 2.29          & 0.82          &  & 2.70          & 0.78          &  & 2.92          & 0.77          &  & \textbf{5.03} & \underline{0.68}          &  & \underline{24.03}          & \underline{0.43}          &  & \underline{49.45}             & \underline{0.32}           \\
GAU-α   w/ I.S.          & \textbf{2.16} & \underline{0.84}          &  & \textbf{2.48} & \textbf{0.81} &  & \textbf{2.71} & \textbf{0.78} &  & 8.60          & 0.58          &  & 109.15         & 0.23          &  & \textgreater{}500 & \textless{}0.1 \\
GAU-α   w/ C.S., I.S.    & 2.28          & 0.82          &  & 2.64          & 0.79          &  & \underline{2.86}          & \underline{0.78}          &  & \underline{5.03}          & \textbf{0.68} &  & \textbf{23.24} & \textbf{0.44} &  & \textbf{44.07}    & \textbf{0.34} 
\\
\bottomrule
\end{tabular}
\end{table*}

\begin{figure*}[t]
\vskip 0.2in
\begin{center}
\centerline{\includegraphics[width=\textwidth]{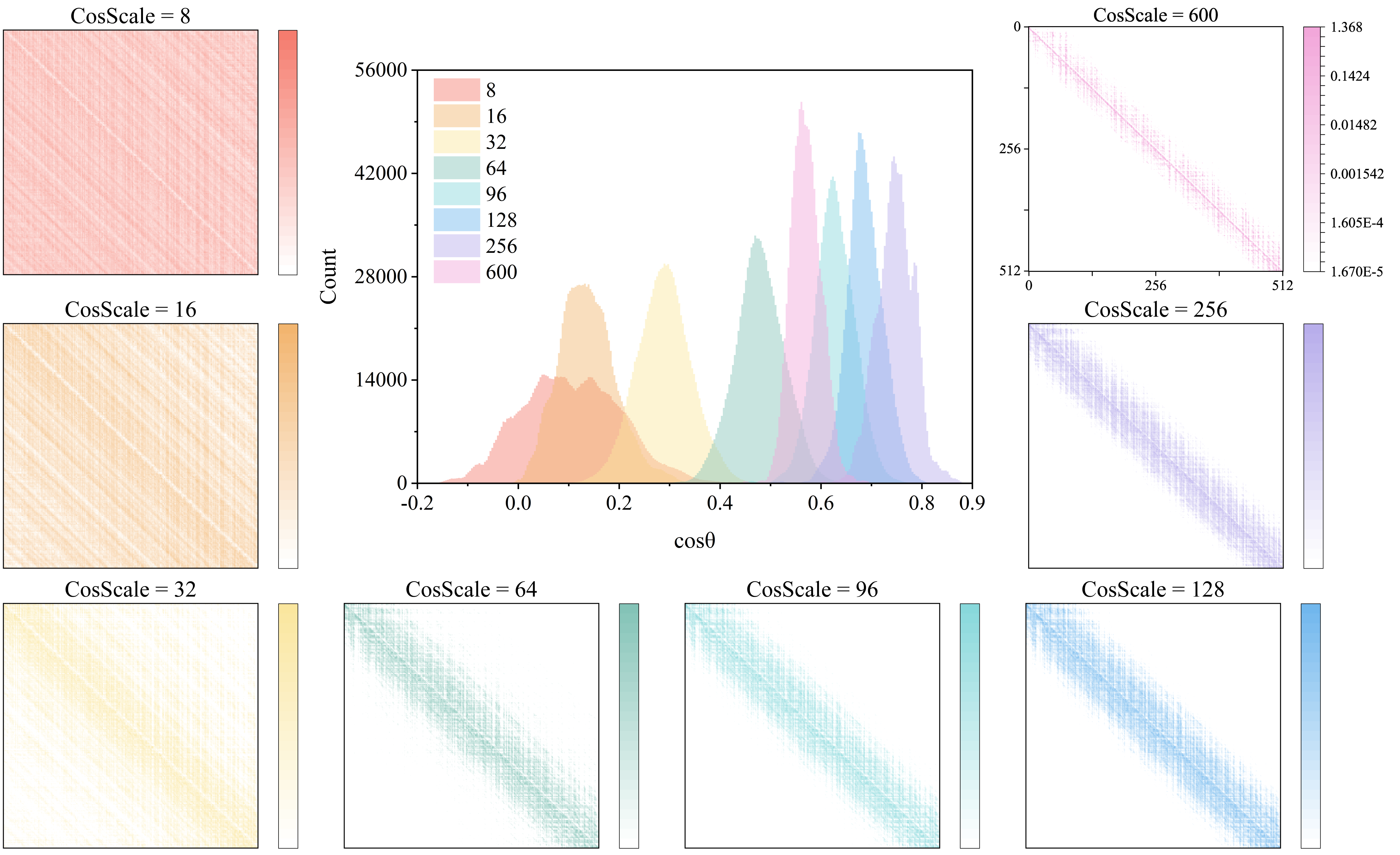}}
\caption{Angle histograms of the attention matrix (after softmax) in the final layer and corresponding heatmaps of the attention matrix (QK) for six-layer models trained with different CosScale values, extrapolated to a sequence length of 512. To better display the diagnal in the heatmap, we apply a natural logarithm transformation to the values, and all heatmaps share the same scale.}
\label{f1}
\end{center}
\vskip -0.2in
\end{figure*}

\begin{figure*}[t]
\vskip 0.1in
\begin{center}
\centerline{\includegraphics[width=\textwidth]{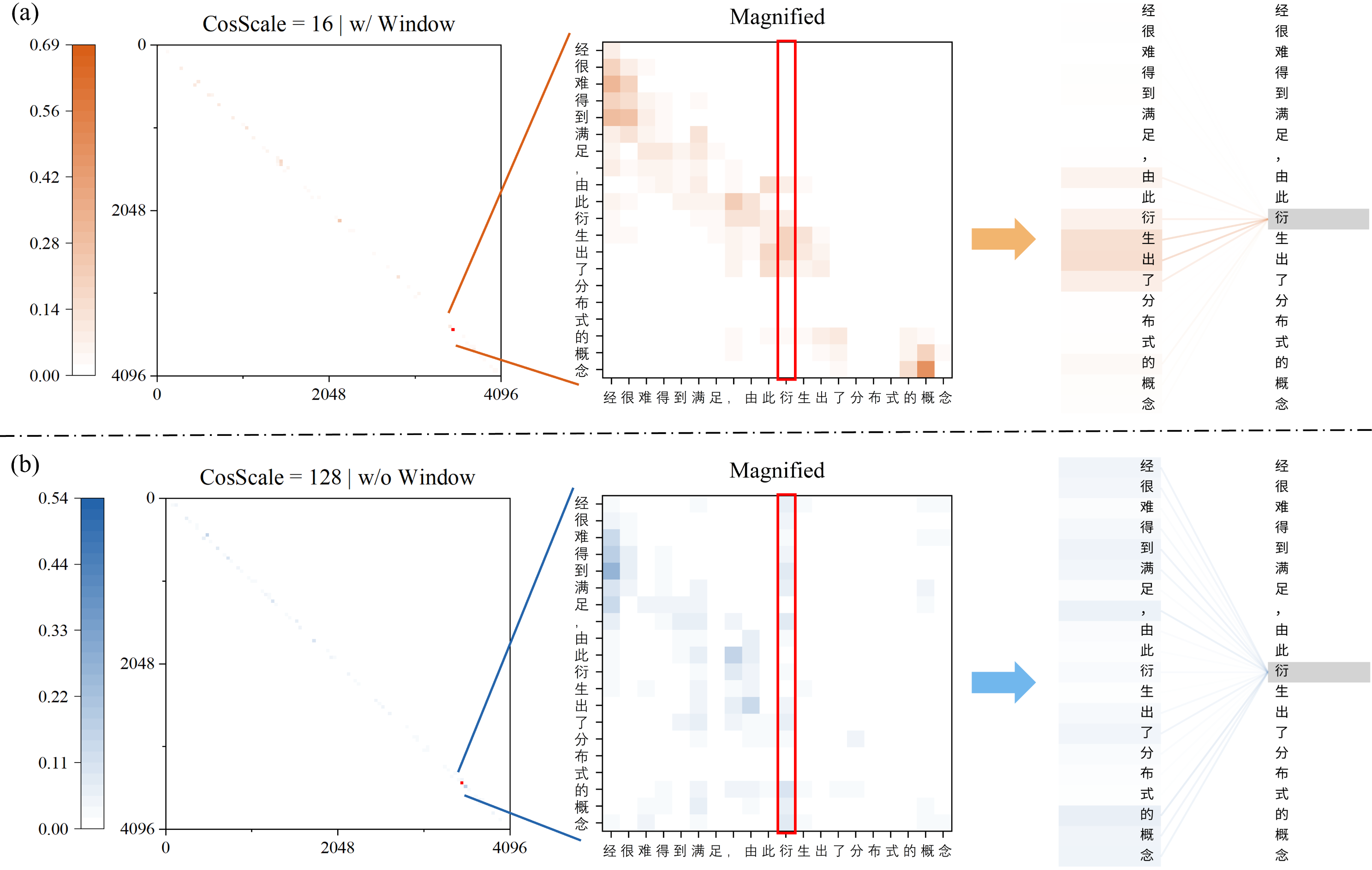}}
\caption{Heatmaps of attention matrices extrapolated to 4096. Panel (a) shows GAU-α with CosScale = 16 applied to the Windowed Attention, while panel (b) presents GAU-α with CosScale = 128 applied to cosine attention. The x- and y-axes represent token positions. A 20x20 local region (from index 3516 to 3535) is extracted from the left heatmap in panel (a) for detailed visualization. The color bar of the left heatmap is scaled to match the value range of the extracted region. }
\label{f2}
\end{center}
\vskip -0.1in
\end{figure*}

\end{document}